\documentclass[preprint,12pt]{elsarticle}

\usepackage{amssymb}
\usepackage{float}
\usepackage{placeins} 

\usepackage{graphicx}
\usepackage{subcaption}

\usepackage{hyperref}
\usepackage{url}
\usepackage{natbib}

\journal{}

\begin{document}

\begin{frontmatter}



\title{Aerial Flood Scene Classification Using Fine-Tuned Attention-based Architecture for Flood-Prone Countries in South Asia}

\author[inst1]{Ibne Hassan}
\ead{ibne.hassan@g.bracu.ac.bd}

\affiliation[inst1]{organization={Computing for Sustainability and Social Good (C2SG) Research Group, Department of Computer Science and Engineering, School of Data and Sciences, BRAC University},
            city={Dhaka},
            country={Bangladesh}}

\author[inst1]{Aman Mujahid}
\ead{aman.mujahid@g.bracu.ac.bd}
\author[inst1]{Abdullah Al Hasib}
\ead{abdullah.al.hasib@g.bracu.ac.bd}
\author[inst1]{Andalib Rahman Shagoto}
\ead{andalib.rahman.shagoto@g.bracu.ac.bd}
\author[inst1]{Joyanta Jyoti Mondal}
\ead{joyanta.jyoti.mondal@g.bracu.ac.bd}
\author[inst1]{Meem Arafat Manab}
\ead{meem.arafat@bracu.ac.bd}
\author[inst1]{Jannatun Noor}
\ead{jannatun.noor@bracu.ac.bd}

\cortext[cor1]{Corresponding author : Aman Mujahid}

\begin{abstract}

Countries in South Asia experience many catastrophic flooding events regularly. It takes time to execute Search and Rescue (SAR) missions in such flooded areas. With the help of image classification, it is possible to expedite such initiatives by classifying flood zones and other locations of interest like houses and humans within such regions. In this paper, we propose a new dataset to enhance SAR by collecting various aerial imagery of flooding events across South Asian countries. For the classification, we propose a fine-tuned Compact Convolutional Transformer (CCT) based approach and some other cutting-edge transformer-based and Convolutional Neural Network-based architectures (CNN). We also implement the YOLOv8 object detection model and detect houses and humans within the imagery of our proposed dataset, and then compare the performance with our classification-based approach. Since the countries in South Asia have similar topography, housing structure, the color of flood water, and vegetation, this work can be more applicable to such a region as opposed to the rest of the world. The images are divided evenly into four classes: `flood', `flood with domicile', `flood with humans', and `no flood'. After experimenting with our proposed dataset on our fine-tuned CCT model, which has a comparatively lower number of weight parameters than many other transformer-based architectures designed for computer vision, it exhibits an accuracy and macro average precision of 98.62\% and 98.50\%. The other transformer-based architectures that we implement are the Vision Transformer (ViT), Swin Transformer, and External Attention Transformer (EANet), which give an accuracy of 88.66\%, 84.74\%, and 66.56\% respectively. We also implement DCECNN (Deep Custom Ensembled Convolutional Neural Network), which is a custom ensemble model that we create by combining MobileNet, InceptionV3, and EfficientNetB0, and we obtain an accuracy of 98.78\%. The architectures we implement are fine-tuned to achieve optimal performance on our dataset. 
\end{abstract}


\end{frontmatter}

\section{Introduction}

Floods are considered to be among the most severely impactful and frequent natural disasters in the world \cite{report_2020-tn}. Due to a large number of countries consisting of coastal regions, the impact of floods on the human establishment and lifestyle is severe, especially for countries in South Asia like Bangladesh, India, and Pakistan, where human casualties due to floods are among the highest \cite{Azeem2023-xt}. A big portion of landmass below sea level, high precipitation, rise in sea level due to climate change, and many other factors contribute to this disaster that endangers millions of people in this region by destroying their homes, resulting in the death of thousands and the displacement of millions of people \cite{Azeem2023-xt}.

For example, the flood that took place in the northeast region of Bangladesh recently in June of 2024, has left around 1.8 million people stranded, as the flood rendered many homes submerged in water \cite{Regan2024-ka}. During such flooding events, the majority of the people are either unprepared or unable to take preventative measures. People who live in coastal areas of Bangladesh experience extreme flooding every year, which results in heavy casualties and housing infrastructure damages \cite{ifrcMillionsBangladesh}. A massive flood that took place in Pakistan in 2022 has placed one-third of the country under water, affecting around 33 million people in the process \cite{unicef_pakistan-ly}. Usually, government agents and other aid agencies working in those areas use boats and aircraft to physically search for survivors which kills ample time lowering the rate of immediate help procedures.

For decades, the flood crisis in South Asian countries has been escalating every year, as the number of deaths, the number of people rendered homeless and the magnitude of the damages keep rising. From being identified as a national crisis by the government, this is now seen as a global crisis with many countries around the world actively participating in providing aid and solutions to reduce the casualties this disaster brings to so many people in in this region. When it comes to short-term responses, providing food and shelter has always been the priority to prevent casualties. But to carry out short-term responses the survivors need to be located first, as houses and landmarks get obscured from ground level.

A similar work to tackle the post-flood disaster was done in Pakistan by Munawar et al. \cite{munawar2021uavs} where they use Unmanned Aerial Vehicles (UAV) to identify flood-related disaster by implementing the Convolutional Neural Network (CNN). Another study \cite{9460988} uses the dataset of Hurricane Harvey that occurred in Texas, United States of America. The authors use models of CNN to determine the post-flood-related catastrophe around the region. Iqbal et al. \cite{iqbal2022floodborne} use object detection to detect flood-borne objects that are responsible for blocking the escape routes of flood water. Besides, Jackson et al. \cite{jackson2023flood} use various CNN models for flood image classification using the FloodNet Dataset. Many of such previous studies have encouraged us to carry out this research work to come up with solutions to the flood crisis in this region.

The primary focus for the development of our work is to help reduce the number of casualties that are likely to occur during the flood crisis in South Asia. To do so, rescue operations need to be carried through faster and more accurate proceedings, which can be done through UAV deployments that will help to identify and eventually map the locations of flooded zones.

The importance of improving the search and rescue initiative is paramount in South Asian countries, as looking for survivors in a harsh climate by the likes of this region makes the initiative a heavily complicated task itself. By contributing to this initiative, we aim to help the concerned authorities expedite the entire system of search and rescue via UAVs through the utilization of image classification. Through classification, it will be much easier for the concerned authorities to map the presence of flooded areas and houses and humans within these areas. With its introduction, Vision Transformers are now widely used for the execution of such image classification techniques. Since our classification will have to be implemented through a mobile component like drones or aircraft, lower computational resources will be required, for which we implement Compact Convolutional Transformers (CCT) in our work. The following are the contributions that we make in this study:

\begin{itemize}

  \item We propose a new dataset consisting of aerial imagery of various flooding events in South Asia that includes four categories: flood, flood with domicile, flood with humans, and no flood. We are specifically focusing on South Asian flood-prone countries, which has not been done before as most of the aerial flood image datasets include imagery focused on other regions of the world.

  \item We experiment on our dataset by implementing a fine-tuned Compact Convolutional Transformer (CCT) and some other cutting-edge transformer-based architectures and CNN-based architectures used in computer vision. We compare the accuracy, precision, recall, F1-score and Matthews correlation coefficient (MCC) obtained from each of the implementations. We also implement all these models in a different aerial flood image dataset and compare the results with our dataset.

  \item We implement YOLOv8, which is an object detection model, to detect houses and humans within the imagery of our dataset, and then analyze the differences between classification and detection in this particular scenario.

   \item We experiment on another aerial flood imagery dataset, known as the FloodNet dataset \cite{9460988}, by applying the same architectures in a similar approach, after which we analyze and compare the results of both implementations.

\end{itemize}

We have divided the rest of the paper into the following sections: in Section \ref{Related_Work}, we review various studies that are related to our field of research. After that, in Section \ref{Background}, we discuss the various architectures that we implement by giving a brief overview of each of their working mechanisms. In Section \ref{Dataset}, we extensively discuss the process of creating our proposed dataset. Then, in Section \ref{Methodology}, we discuss manual parameters for each architecture and the tuning approach that we implement for each of them. After that, in Section \ref{Experimental_Setup}, we showcase the experimental setup and experimental results, then compare the implemented approaches based on the obtained results. We discuss the real-world applicability of our implemented approaches in Section \ref{Discussion}. Before concluding our paper, we discuss the limitations and any future implementations that we plan to execute in Section \ref{Limitations and future work}.

\section{Related Work}
\label{Related_Work}

For our research work, we are implementing flood zone, human, and house classification by processing images from our proposed dataset. To do so, we take inspiration from the various related works that have been published in recent years.

\subsection{Flood Detection and Mapping}

To carry out search and rescue, the flood-inundated areas need to be identified by our UAV first. This can be done today through the implementation of various machine learning based models \cite{10156673, 9800023}. But, putting aside the use of image processing techniques and aerial imagery that we are proposing in this paper, flood mapping was and is still quite popularly done using the Geographical Information System (GIS) creating a digital elevation model based on topographical data using software like ArcGIS \cite{biswas2018methodologies}. We can observe that such GIS-based methods help evacuate people from certain regions based on the calculated risks of that particular region. However, when it comes to carrying out search and rescue from already flooded areas, a different approach of flood mapping is required to simultaneously detect the flooded area and carry out rescue operations. Now, in terms of flood detection, Remote Sensing (RS) based methods are also very popular in terms of obtaining data. Here, the technologies based on RS that are used are Light Detection and Ranging (LIDAR), Multispectral imaging via satellites, Radar, etc which provide data that can be interpreted as flood-prone areas or flooded zones \cite{munawar2022remote}. 

As we are proposing an image processing-based approach, unlike GIS-based methods, we can classify and eventually map flood zones during and after the occurrences of floods. We will not be requiring multiple technologies like LIDAR, Satellite Imagery, etc. as we can just classify flood zones within aerial imagery from UAVs. When it comes to flood detection and mapping, various papers have aimed to achieve this feat via image processing techniques, especially with the help of neural network-based models. For example, Rizk et al. \cite{rizk2022drone} have opted to use the VGG-16 network to get an estimate of the water level around objects like houses, cars, etc.

\subsection{House and Human Detection}

Another aspect of digital image processing is house detection or any other infrastructure from a given height mainly through satellite or aerial photography. Quite similar to the human detection process, the high-resolution RGB images taken for houses are differentiated from the green landscape to carry out object detection through image processing techniques. Both Lygouras et al. \cite{lygouras2019unsupervised} and Liu et al. \cite{liu2021real} use UAVs to detect human bodies. but this time the recognition made by the algorithm is unsupervised. In the case of \cite{lygouras2019unsupervised}, they have used an algorithm which makes unsupervised recognition detecting endangered swimmers, with the help of hexacopters, as part of the SAR operation. They also used the YOLO architecture to carry out human detection from datasets that were collected from videos made by the research members. The challenge they face is the outcome of the silhouette of the human body that is apparent on the water's surface. The algorithm was able to detect human beings with an accuracy of about 67\%.

\subsection{Image Classification and Transformers}

Here, we discuss various research papers that have leveraged the use of image classification using various models including transformer-based architectures, which are relative to our work. Classifying floods through CNN models has become a demanding approach for many developing countries which are prone to seasonal floods every year. Roy et al. use Flood Transformer \cite{roy2022transformer} which is a hybrid of transformer and CNN model accompanied by Flood Capacity metric to also measure the extent of water level from the surface. Researchers have used several models on the FloodNet dataset such as ResNet-18 \cite{jackson2023flood}, UNet-MobileNetV3 and PSPNet \cite{safavi2021comparative}, DeepLabv3 and Segformer \cite{gupta13post}. All of these are based on the state-of-the-art (SOTA) semantic segmentation model \cite{mo2022review}. We also reviewed some works that focus on the urban environment using Vision Transformer \cite{le2023fl} to measure the water level and also Faster R-CNN \cite{iqbal2022floodborne} to detect any flood-borne objects. Seyed et al. used PSPNet, TransUNet, and SegFormer \cite{erfani2023eye} models on time-lapse images to measure water levels to enhance flood detection. All of these papers are successful in acquiring an accuracy of more than 80\% but they are restricted to binary classification, i.e. `flood' and `no-flood'.

Combining a transformer model and a conventional convolutional algorithm has advantages in terms of parameter efficiency and the ability to process long-range and global dependencies or interactions between various parts of an image. The architecture of the CCT provides the foundation where the model needs fewer parameters, as suggested by Hassani et al \cite{hassani2022escaping}. This model stands out because it eliminates boundary-level data that is present between several patches. This gives neural networks the freedom to investigate different complexities. One of the notable works is produced by Jajja et al \cite{jajja2022compact}, where they emphasize the CCT model over MobileNet, ResNet152v2, and VGG-16 on preventing cotton pest attacks due to obtaining the highest accuracy. Khan et al. \cite{khan2023computer} use a fine-tuned CCT model to carry out Diabetic Retinopathy (DR) classification among the victim patients. Even with low-pixel images from the datasets used, the CCT model gives a test accuracy of 90.17\%. CCT's high accuracy in medical research allows \cite{sun2022cct} to carry out a lung disease classification using a dataset that consists of chest images from the Computed Tomography (CT) results. This time the model achieves an accuracy of 98.6\%.

Munawar et al. \cite{munawar2021application} takes an approach of flood mapping solely based on CNN by passing images of flood zones through the many layers of the CNN architecture. They obtain an overall accuracy of 91\% in terms of classifying flood zones. Munawar et al. \cite{munawar2021uavs} have also opted for the use of CNN in another work by training the model through trimming pre-disaster and post-disaster images into patches and carrying out comparisons, which results in a huge number of iterations. The research work done by Afridi et al. \cite{afridi2019flood} has proposed an approach to detecting water in flooded zones based on thermal imaging. Here, the Hue Saturation Value (HSV) color model is used to identify water bodies based on the color spectrum of water. Many of the other works that we reviewed have also managed to obtain accuracies ranging from 70\% to 90\%, and many even beyond the 90\% benchmark, thus proving the flood detection and mapping methods to be quite successful.

In the research paper demonstrated by Hernández et al., semantic segmentation is carried out on the imagery, where each pixel of the image is classified based on categories to carry out prediction at pixel level \cite{hernandez2022flood}. Another approach to flood mapping has been proposed by Gebrehiwot et al. \cite{gebrehiwot2019deep}, where the FCN-16s model is used. A Fully Convolutional Network (FCN) structure has been used because, unlike the CNN-based VGG-16 network, FCN can provide a 2-dimensional class map where VGG-16 can provide a 1-dimensional class map. There were four output classes of FCN-16s, which were water, road, vegetation, and building, based on which the mapping was provided as the result.

\subsection{Implementation on Mobile Device}

The application of Transformer and CNN-based models on mobile devices is a convenient method for image classification. Drones are commonly used to carry out research in isolated parts of the world. Jamil et al. \cite{jamil2023comprehensive} shows a comprehensive survey on how drones are used to take aerial images to classify specific elements of research using various transformer and neural network-based models. High-resolution images are taken by drones or mobile devices which are then processed by the models that are installed within the device. As mentioned in the paper, drones can reduce anomalies in images affected by clouds or trees as opposed to traditional methods like taking pictures from helicopters or airplanes. Gibril et al. \cite{gibril2023large} use drones to identify date palm trees from the pictures taken. Variants of vision transformer models such as the Segformer, Segmenter, Upper-Net Swin Transformer, and dense prediction Transformer are used to evaluate the palm trees. The accuracy of the models ranged from 91.62\% to 92.44\%.

Most of the research works that we study implement flood scene classification based on classifying only the flooded zones within their images, while we propose an approach to classify humans and houses within the flood zones as well. In our study, we solely focus on South Asian flood imagery, which is not the case for many of the recent works as they either focus on a specific country or just worldwide imagery altogether.

\section{Background}
\label{Background}

In this section, we discuss the transformer-based architectures and the CNN architectures that we implement both individually and as a custom ensemble model.

\subsection{CNN-based Architectures}

When it comes to neural network architectures, the focus is usually to obtain as much accuracy as possible. But this is not the only factor that researchers try to keep in mind when it comes to practical applications, like self-driving cars and robotics. In this case, the factor of computational efficiency comes into play as these factors determine how these architectures can carry out such image-processing tasks on limited computational resources. For this, Howard et al. \cite{howard2017mobilenets} present an effective neural network design known as MobileNet, which can be utilized for object recognition, detection, and classification in embedded vision applications.
 To build the lightweight neural network, it implements the concept of depth-wise separable convolutions. For embedded applications, combating latency is vital while also maintaining good accuracy, for which Howard et al. \cite{howard2017mobilenets} introduce two global hyperparameters which let us choose the right size of the model. This model works well with embedded applications, and it aligns with our work since we are working with aerial imagery from UAVs and aircraft.

To counter computational complexity in the CNN architectures, the Inception model was introduced by the paper \cite{szegedy2014going} which simultaneously performs multiple convolutions with different filter sizes and pooling operations. But it was difficult to bring changes in the network architecture as scaling it up will result in large parts of its computational gain being lost \cite{szegedy2015rethinking}. To further improve the computational efficiency by carrying out the scaling efficiently and also improve performance, the same authors propose an enhanced version of this model known as InceptionV3 \cite{szegedy2015rethinking}. The Inception model and its enhanced versions like the InceptionV3 model are highly regarded for image classification tasks, which is why we implement this architecture on our dataset to showcase how well image classification tasks can be carried out within our imagery.

A different approach in scaling up ConvNets is proposed by Tan et al.\cite{tan2020efficientnet} at the same time maintaining better efficiency and accuracy, known as EfficientNetB0. Through a fixed set of scaling coefficients, the method this paper proposes carries out uniform scaling of the network's resolution, width, and depth, which the paper refers to as the compound scaling method. This method is more advantageous in maintaining accuracy as it tries to maintain a balance between depth, width, and resolution instead of arbitrarily scaling them \cite{tan2020efficientnet}. This balanced scalability allows us to choose the right-sized model for our specific tasks and resource limitations, for which this architecture is a proper fit for our work.

\subsection{Vision Transformers (ViT)}

Helping to bring the Transformers architecture to the world of computer vision, Vision Transformers \cite{dosovitskiy2021image} has proved to be one of the high-performing models of recent times. In comparison to traditional models used in computer vision like CNN, image classification can be carried out by applying transformers on sequences of image patches instead of applying attention to convolutional networks and therefore relying on it. Thus, it requires less computational resources and therefore performs very efficiently compared to convolutional networks. The process of Vision Transformers involves dividing an image into patches, then embedding each patch linearly, including positional embeddings, and finally feeding the resulting sequence of vectors to a Transformer encoder. A learnable classification token is added to this sequence to carry out image classification, which aligns with the primary objective of our work \cite{dosovitskiy2021image}.

\begin{figure}[!ht]
\centerline{\includegraphics[width=5.5in, height=3.5in]{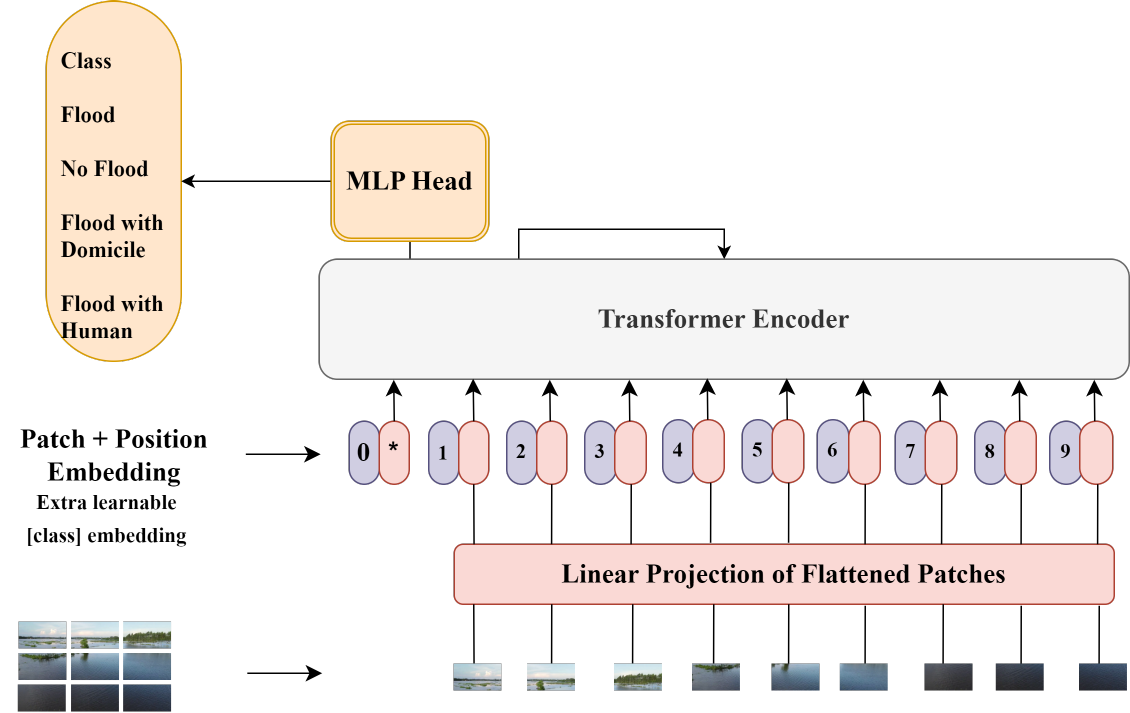}}
\caption{Vision Transformers Architecture}
\label{vitarc}
\end{figure}

As this is a transformer-based architecture designed for computer vision, it requires an image input instead of textual input. To take image input, the image is divided into patches of fixed size and needs to be non-overlapping. The linear patch embedding method is conducted by flattening the patches before embedding them. The embedding process is similar to the word embedding process that is carried out in the standard  Transformers model that is implemented for natural language processing. The spatial information for these patch embeddings is also required, for which, positional encodings are added to them to know the relative position of each of the patches. Figure \ref{vitarc} illustrates the Vision Transformer architecture that we implement in our work.

\subsection{Swin Transformer}

Even though Vision Transformers (ViT) has been and is still a state-of-the-art architecture to carry out computer vision tasks, there are still some drawbacks which include its computational capability being quadratic to the size of the image \cite{Loy2022-he}. Thus, this results in increased difficulty for the ViT model to work with images of such high resolution and even the fixed scale tokens do not suit well when there are images of variable size \cite{Loy2022-he}.

Swin Transformer helps to enhance the capability of the Transformer to deal with vision tasks of higher complexity, for instance, high-resolution pictures, with the help of the Shifting Windows \cite{Liu_2021_ICCV} scheme to form a hierarchical structure. After an RGB image is split into patches which are then treated as tokens, several Swin Transformer Blocks (blocks with modified self-attention computation) are applied on these patch tokens after its raw-valued feature has a linear embedding layer applied \cite{Liu_2021_ICCV}. 

\subsection{Compact Convolutional Transformer (CCT)}

Usually, Transformer-based architectures that are applied for computer vision, like Vision Transformers (ViT) mentioned above, require a very large dataset. However, this is usually not an issue when implementing CNN-based models as they have quite well-informed inductive biases due to the usage of convolution. So, to get the best of both worlds, the Compact Convolutional Transformer (CCT) model is introduced.

\begin{figure}[!ht]
\centerline{\includegraphics[scale=.5]{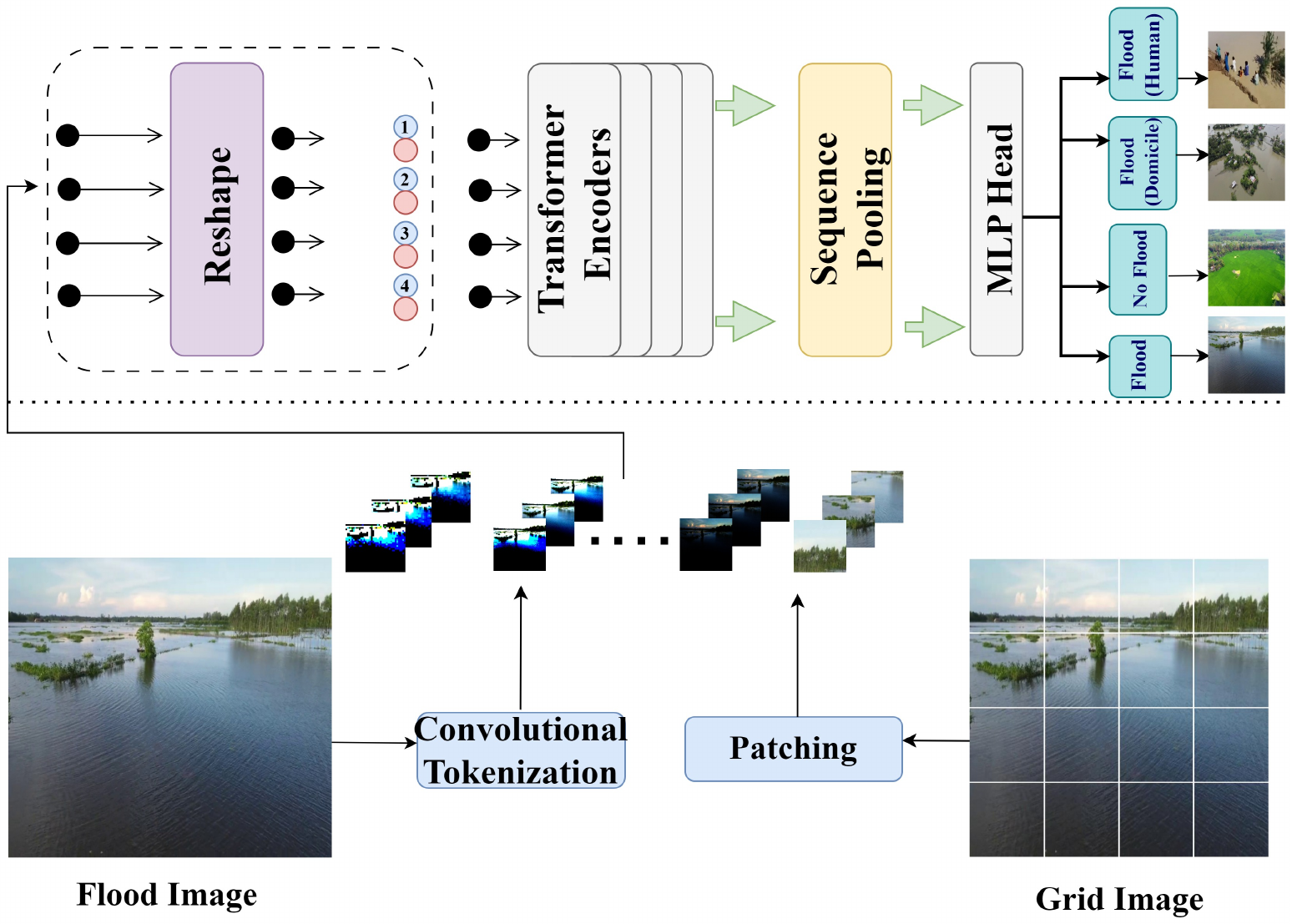}}
\caption{Compact Convolutional Transformer (CCT) Architecture}
\label{cct}
\end{figure}

The development of Compact Convolutional Transformer \cite{hassani2022escaping} is another breakthrough in enhancing the processing of image, as it gets the upper hand when competing against Vision Transformers in terms of being more compatible in working with smaller datasets. While small in size and compact, this model can still perform quite well as it can be quickly trained from scratch by a new and small dataset \cite{hassani2022escaping}. This level of accessibility in the model is what aligns it perfectly with the scale of our work and our dataset, as we can maintain its authenticity by augmenting a bit less compared to other models and still obtain adequate accuracy. CCT is formed by introducing a sequence pooling and then convolutional blocks are added to the tokenization \cite{hassani2022escaping}. Figure \ref{cct} below describes the process by displaying the overall architecture of this model.

\subsection{External Attention Transformer (EANet)}

Deep feature representation for visual tasks has become increasingly dependent on attention processes, particularly self-attention. To capture the long-range dependency inside a single sample, self-attention computes a weighted sum of features utilizing the pair-wise affinity's overall positions. Self-attention, however, ignores the possibility of sample correlation and has quadratic complexity \cite{guo2021selfattention}. Self-attention refines each position's representation by combining information from all other locations in a single sample, which results in quadratic computational complexity as the number of sites in a sample increases \cite{guo2021selfattention}.

EANet introduces the concept of external attention \cite{guo2021selfattention}, which is a new attention mechanism that uses two normalization and two cascaded layers that help to implement two external, learnable, small, and shared memories. It has linear complexity due to considering the sample correlations to be implicit.

\subsection{DCECNN (Deep Custom Ensembled Convolutional Neural Network) model}

Ensemble learning is a method of machine learning that combines multiple models and generates a single architecture that helps to bring the maximum predictive outcome \cite{Lutins2017-rg}. Instead of working on getting the best out of one model, we strategically combine several models and then average them. The accuracy or forecast becomes fully skewed as relying on just one model might lead to excessive bias and poor performance. Thus, by integrating many models and employing a macro average voting approach, high-performing models can contribute more and low-performing models less. Hence, we assemble three CNN architectures to form the DCECNN (Deep Custom Ensembled Convolutional Neural Network) model, which combines the best attributes of our three chosen architectures to provide optimum performance.

\begin{figure}[ht]
\centerline{\includegraphics[scale=.7]{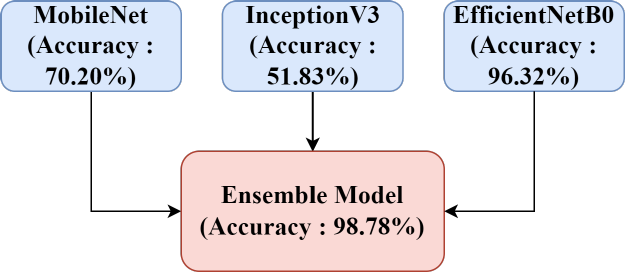}}
\caption{Structure of the DCECNN model and the individual and collective accuracies}
\label{ensemble structure}
\end{figure}

In this research, we merge MobileNet, InceptionV3, and EfficientNetB0, three different pre-trained Convolutional Neural Network (CNN) based architectures. Based on a streamlined architecture, MobileNets use depth-wise separable convolutions to build lightweight deep neural networks \cite{howard2017mobilenets}. The biggest advantage of using MobileNet that aligns well with our work is that it requires less computation as it uses a small network and therefore can work well with resource restrictions \cite{howard2017mobilenets}, for example, in mobile devices like drones. But then again, due to lower complexity and network size, it also gives lower accuracy compared to many other complex models like Inception. So, to counter this drawback, we use InceptionV3 \cite{szegedy2015rethinking} as part of our ensemble learning. InceptionV3 is a deep convolutional neural network that was a rethinking of the previous Inception architecture by improving computational efficiency and better accuracy as a much deeper network is used \cite{szegedy2015rethinking}. Another model that we implement as part of our ensemble learning method is the EfficientNetB0 \cite{tan2020efficientnet}, which is known to have higher scalability compared to that of InceptionV3 and MobileNet and also better transfer learning capability \cite{tan2020efficientnet}. The three models that we implement help to counter the drawbacks of each other and therefore provide optimum accuracy. The structure of our custom ensemble model DCECNN is summarized in Figure \ref{ensemble structure}.

\subsection{YOLOv8}

The YOLOv8 model is among the latest state-of-the-art YOLO models that are widely regarded for object detection or image segmentation applications. Developed by Ultralytics, this model is pre-trained with the COCO dataset, and the version of YOLOv8 that we implement, which is the YOLOv8m, has 25.9 million parameters \cite{Ultralytics_undated-xz}. This is approximately 65 times higher than the number of total parameters in CCT and 118 times higher than the number of total parameters in Swin Transformer. An official paper regarding YOLOv8 is yet to be released. Since our work at its core involves the identification of flood zones and other locations of interest, the use of detection models in such types of scenarios is quite common. However, in our work, we are focusing on executing image classification instead of detection by leveraging the use of low-parameter transformer models as classification itself generally requires fewer computational resources than detection. Aside from image classification having the upper hand in resource efficiency, the performance delivered also stands to be an important factor as the culmination of both good performance and efficiency is needed for a model or a technique to be considered a far better approach than another one. So, we implement the YOLOv8 model with our dataset to carry out object detection and observe the performance it delivers on our proposed dataset.

\section{Dataset}\label{Dataset}

In this part, we present the approaches that we undertake to create our proposed dataset, which includes the formation of the raw dataset and the methods to carry out preprocessing.

\subsection{Dataset Description}

\begin{figure}[htbp]
\centerline{\includegraphics[width=5in, height=1.5in]{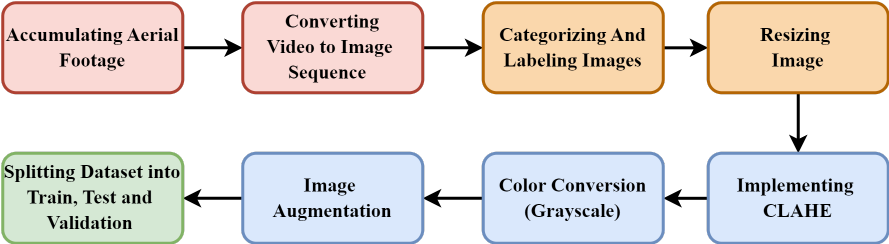}}
\caption{Workflow of our dataset preparation.}
\label{fig: workflow}
\end{figure}

To obtain enough aerial footage of flooding events that have taken place within the most flood-prone countries in South Asia, we create the AFSSA (Aerial Flood Scene South Asia) dataset. Most of the flood-related datasets available include flood imagery from all around the world and some of them include images that are either beyond aerial, like geospatial imagery, or from angles that may not represent the sort of footage that is captured from aerial vehicles like drones and helicopters. Since we focus on imagery that is captured on drones or aerial vehicles, our best course of action is to obtain such imagery from actual drone videos taken during flooding events within our targeted region. The reason to work only on such aerial imagery is that our work is targeted to help the search and rescue or relief teams via UAV implementation. Secondly, we chose a specific region like South Asia over the rest of the world because the countries in this region have a specific terrain, housing structure, human posture, color of flood water, and vegetation, which makes it much easier to train the models and for the models to classify. For example, the housing structures in flood-prone areas within South Asia are quite different from other parts of the world as most of the houses are built with different materials. For instance, the use of tin sheds in houses makes them look completely different from aerial view as opposed to houses in western parts of the world like the US or Canada. For this reason, we propose a dataset that would specifically consist of aerial footage within the flood zones of countries within South Asia. Figure \ref{fig: workflow} shows the workflow that denotes the approaches we took to prepare our dataset.

\subsection{Data Collection}

To create the AFSSA dataset, we surf through YouTube for footage of flood events in this region. We obtain various drone videos from countries including Bangladesh, India, Pakistan, and Indonesia. These sources ranged from individual drone users to established news channels, providing a comprehensive collection of aerial flood footage.

The YouTube user and channel Hannan Miah uploaded several drone videos \cite{Hannan_Miah} of the recent 2022 flooding in Sylhet, Bangladesh. Similarly, the YouTube channel AZ UK provided footage \cite{AZ_UK} of the same event in Sylhet. Desi FireFly, another YouTube channel, shared drone footage \cite{Desi_FireFly} from a different flooding event in Bogra, Bangladesh, characterized by reddish floodwater. Al Jazeera also featured drone footage of flooding in Bangladesh in one of their news segments \cite{Al_Jazeera_English}. Beyond Bangladesh, we found flood footage from various locations in India, including Assam, Mayapur, and Silchar, uploaded by YouTube channels NIMAINITAI \cite{NIMAINITAI}, just as it is \cite{JustAs_it_is}, DRONE PHOTOGRAPHY, SILCHAR \cite{DRONE_PHOTOGRAPHY_SILCHAR}, Minar Dev \cite{Minar_Dev}, and NH9 News \cite{NH9_News}. Additionally, the YouTube channel Sanaullah Janweri \cite{Sanaullah_Janweri} provided news footage from The Guardian about a flooding event in Pakistan. Several YouTube channels, including Voice of America \cite{Voice_of_America}, Guardian News \cite{Guardian_News}, and TimesLIVE Video \cite{TimesLIVE_Video}, uploaded footage of the aftermath of a tsunami in Indonesia. For aerial footage of non-flooded areas, we collected videos from YouTube channels UJJWAL MISTRI \cite{UJJWAL_MISTRI}, FMI Productions \cite{FMI_Productions}, Izzyvillage \cite{Izzyvillage}, and HIMEL VAI \cite{HIMEL_VAI.}, showcasing drone footage from rural areas in this region.

\subsection{Dataset Formation}

\begin{figure*}[!ht]
    \centering
    \begin{subfigure}{0.24\textwidth} 
        \centering
        \includegraphics[width=\textwidth, height=\textheight, keepaspectratio]{"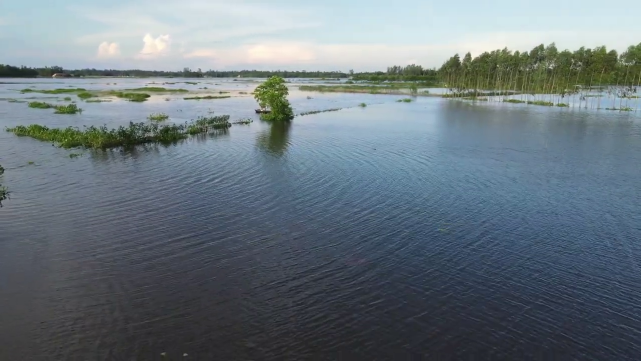"}
        \caption{Flood}
        \label{fig:flood}
    \end{subfigure}
    \hfill
    \begin{subfigure}{0.24\textwidth} 
        \centering
        \includegraphics[width=\textwidth, height=\textheight, keepaspectratio]{"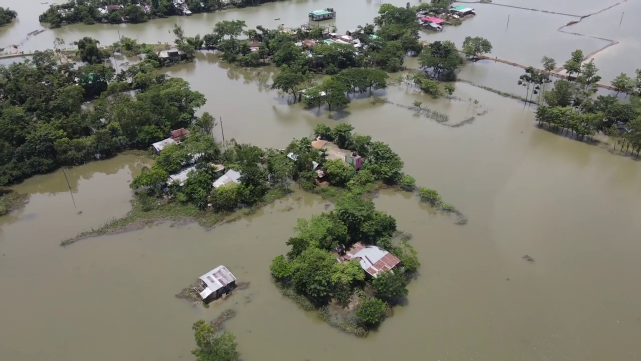"}
        \caption{Flood with domicile}
        \label{fig:domicile}
    \end{subfigure}
    \hfill
    \begin{subfigure}{0.24\textwidth} 
        \centering
        \includegraphics[width=\textwidth, height=\textheight, keepaspectratio]{"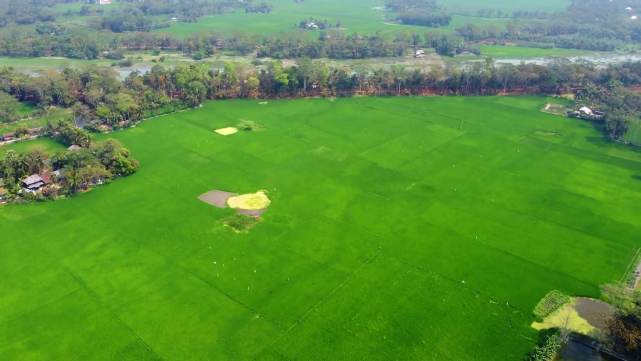"}
        \caption{No flood}
        \label{fig:no_flood} 
    \end{subfigure}
    \hfill
    \begin{subfigure}{0.24\textwidth} 
        \centering
        \includegraphics[width=\textwidth, height=\textheight, keepaspectratio]{"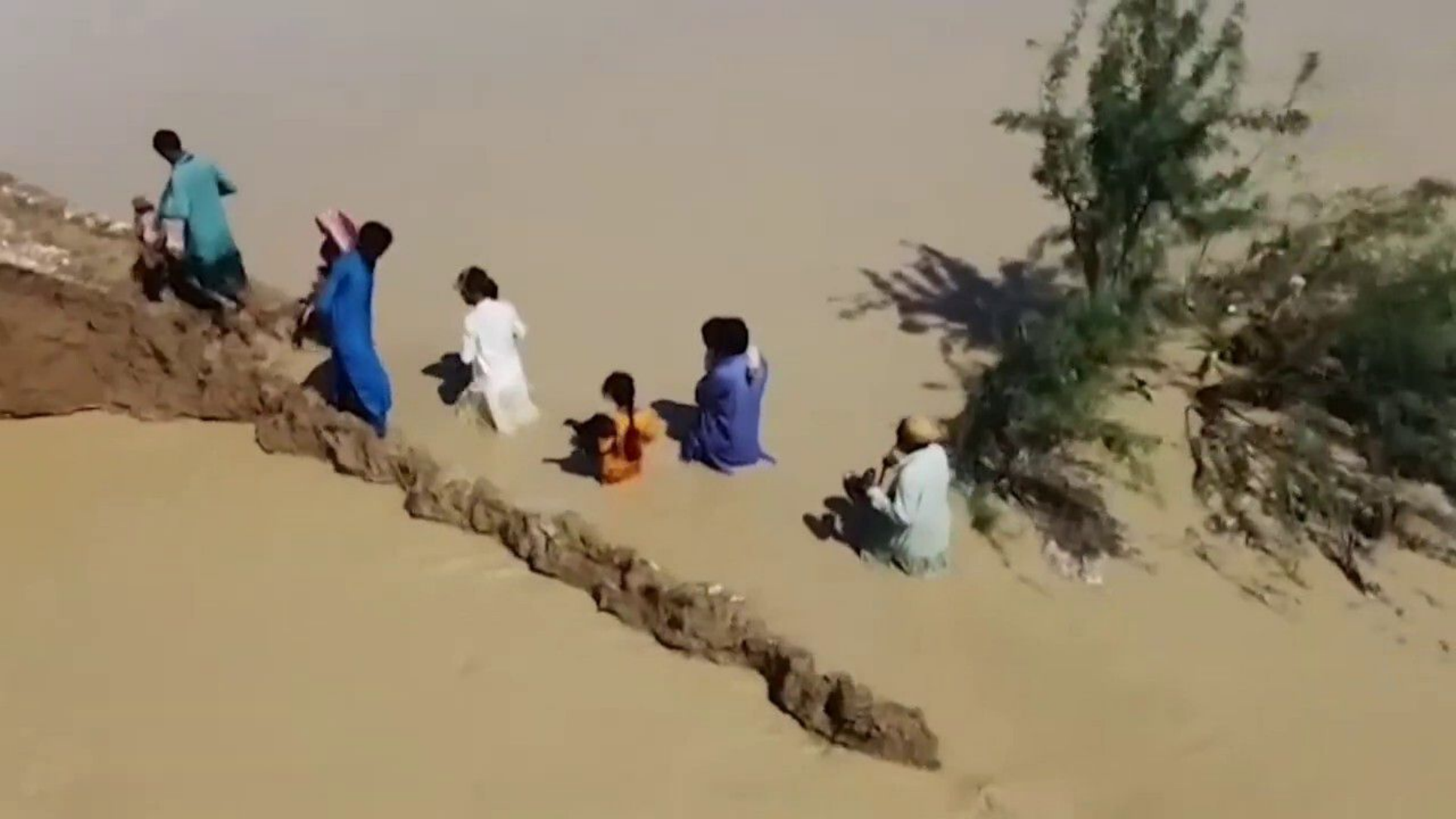"}
        \caption{Flood with humans}
        \label{fig:humans}
    \end{subfigure}
    \caption{Sample image from each category of our proposed dataset.}
    \label{fig:all_category_pics} 
\end{figure*}

After collecting the video footage from all the mentioned sources, we used the Scene Filter option on an open-source video player named VLC Media Player to convert the videos into image sequences. These images are then categorized based on four categories: `flood', `flood with domicile', `flood with humans', and `no flood'. When generating an image sequence, we set the recording ratio as 5. Here, the recording ratio is the frame interval after which a frame is extracted from the video and saved. After obtaining image sequences from the videos, we spread out our selection of images instead of selecting sequentially. For each image selected, four to five images from the continuing sequence are skipped. We do this to increase variation from one image to the next. To reduce biases in our dataset, it is important to maintain variation between the images in this instance since we are extracting images from image sequences of long and continuing drone shots, where several images with very little difference among them are often captured within certain parts of the image sequences. This selection process is followed for each of the four categories. We gather over 300 photos for each category. The number of images in the `flood' category is 305, for the `flood with domicile' category it is 308, for the `flood with humans' category it is 308, and for the `no flood' category it is 307. We try to maintain an approximately equal number of images for each class to form a balanced dataset. Figure \ref{fig:all_category_pics} shows a sample image from each category of the AFSSA dataset.

\subsection{Data Preprocessing}

To expand the AFSSA dataset further, we carry out augmentation to generate more varying images within each category. For dataset augmentation, each image is augmented six times and in six different ways using Keras ImageDataGenerator. Each image is randomly rotated 0$^{\circ}$ to 45$^{\circ}$ with a width shift range of 0.3, height shift range of 0.3, shear range of 0.3 along with horizontal flip, and many other approaches. The total number of images after augmentation is 8603 images in total. The purpose of data augmentation is to increase the number of images in our dataset, as transformer-based architectures require a relatively higher amount of data.

Furthermore, Contrast-Limited Adaptive Histogram Equalization (CLAHE) is used to improve the contrast of the images and to carry out histogram equalization. CLAHE, which is a version of the Adaptive Histogram Equalization (AHE), restricts contrast amplification to lessen noise amplification. We achieve this by successfully extending the intensity range of the picture and spreading out the most common intensity levels. The number of images before and after augmentation is displayed in Table \ref{tab: num} and the approaches taken to carry out the augmentation are also displayed in Table \ref{tab: aug}.

\begin{table}[htbp]
\centering
\small
\begin{tabular}{|c|c|}
\hline
 \textbf{Category} & \textbf{Number of Images} \\
        \hline
         Flood & 305 \\
         \hline
         Flood (Domicile) & 308 \\
         \hline
         Flood with Humans & 308 \\
         \hline
         No Flood & 307 \\
         \hline
         Total & 1228  \\
         \hline
\end{tabular}
\caption{Number of images in the AFSSA dataset for each category.}
\label{tab: num}
\end{table}

\begin{table}[htbp]
\centering
\small
\begin{tabular}{|p{5cm}|p{4cm}|}
    \hline
        \textbf{Augmentation Type} & \textbf{Approach} \\
        \hline
         Rotation ceiling & Randomly from 0 to 45$^{\circ}$ \\
          \hline
         Height shift ceiling & Till 30\% \\
          \hline
         Width shift ceiling & Till 30\% \\
          \hline
         Shear limit & 30\% \\
          \hline
         Zoom limit & 30\% \\
          \hline
         Vertical turnaround & Yes \\
          \hline
         Horizontal turnaround & Yes \\
          \hline
         Staffing function & Reflective \\
         \hline
\end{tabular}
\caption{Approaches taken to augment our dataset.}
\label{tab: aug}
\end{table}

\subsection{Ethical Considerations}

Due to the limited availability of aerial and flooded imagery over the South Asia region, we choose to collect data from videos, selecting a substantial number of frames. YouTube, as the most popular video-sharing platform enriched with diverse scenarios, serves as our primary source for extraction. In collecting images from these videos, we are fully aware of and respectful towards ethical considerations, ensuring we do not misappropriate credit or harm the content owners and organizations. The visibility setting of all the YouTube videos used for this research work is set as public and the videos are not commercially licensed. Content creators or owners of such videos are properly credited by citing the videos appropriately throughout this paper.

\section{Methodology}
\label{Methodology}

The models that we implement consist of a different number of parameters, among these parameters, the number of trainable and non-trainable parameters are also different for each of the models. Table \ref{tab: parameters} shows the number of total, trainable, and non-trainable parameters for each of the models that we implement on our dataset. Here, the trainable parameters represent the variables in our implemented architectures that are adjusted during the training phase. The non-trainable parameters remain constant during the training phase. In Table \ref{tab: parameters}, we can observe that the transformer-based architectures have a higher trainable to non-trainable ratio compared to CNN-based architectures, thus signifying their capability to learn complex patterns better than the CNN-based architectures and be more flexible for fine-tuning. From the table, it is evident that the Swin Transformer model has the lowest number of total parameters, with CCT having the second-lowest.

\begin{table*}[ht]
    \centering
    \resizebox{\textwidth}{!}{
    \begin{tabular}{|c|c|c|c|}
    \hline
    \textbf{Models} & \textbf{Total Parameters} & \textbf{Trainable Parameters} & \textbf{Non-trainable Parameters} \\
    
    \hline
     ViT & 11,211,979 & 11,211,972 & 7  \\
    \hline

    {Swin Transformer} & {222,388} & 217,172 & 5,216  \\
    \hline

    EANet & 310,899 & 310,892 & 7 \\
    \hline

    MobileNet & 3,361,096 & 3,337,992 & 23,104 \\
    \hline

    InceptionV3 & 22,135,208 & 22,072,136 & 63,072 \\
    \hline

    EfficientNetB0 & 4,283,691 & 4,240,388 & 43,303 \\ 
    \hline

    DCECNN & 29,779,995 & 29,650,516 & 129,479\\ 
    
    \hline

    YOLOv8m & 25,900,000 (approx.) \cite{Ultralytics_undated-xz} & - & - \\

    \hline

    \textbf{CCT} & \textbf{407,365} & 407,365 & 0 \\
    \hline
    
    \end{tabular}
    }
    \caption{Total Parameters, Trainable Parameters, and Non-trainable Parameters for each of our implemented models.}
    \label{tab: parameters}
\end{table*}

\subsection{Manual Tuning}

Since we implement multiple architectures in our proposed work, we plan on optimizing the major controllable features of the architectures in a way that can provide us with optimum value for our performance metrics. The parameters that we tune to receive optimum performance metrics are the dropout rate, input size, weight decay, and batch size.

The manual tuning approach allows for a computationally effective tuning approach as it requires less computational resources. This approach allows us to control the tuning process precisely by observing the model's performance at each step and tuning accordingly to maximize its performance. Moreover, This method also aids in improving the understanding of how modifications impact the model's abilities.

We implement each model using a trial-and-error method in which all the parameters are tuned to determine the optimal performance. Table \ref{tab:baseline2} shows the hyperparameters, which are the dropout rate, input size, weight decay, and the batch size of the CCT model of each step of our fine-tuning, with the first step being the top row and going all the way to the bottom row, which is the final tuned setting for the parameters. In Table \ref{tab:baseline2}, the initial results of CCT are 87.27\% accuracy and 89.50\% precision. However, after changing the hyper-parameters seven times, increasing the input size to 128, and decreasing the batch size to 32 to reduce training loss, the accuracy and precision increase to 99.62\% and 98.50\% respectively.

Table \ref{tab:baseline} displays the final parameter tuning for each of the architectures we implement. Just like the step-by-step process shown for CCT, we carry out the same process for each of the architectures and reach a certain value of dropout rate, input size, weight decay, and batch size that provides the highest possible accuracy and precision with our proposed dataset.

\begin{table}[ht]
    \centering
    \resizebox{\textwidth}{!}{
    \begin{tabular}{|c|c|c|c|c|c|}
         \hline
         \textbf{Dropout Rate} & \textbf{Input Size} & \textbf{Weight Decay} & \textbf{Batch Size} & \textbf{Accuracy(\%)} & \textbf{Precision(\%)} \\
         \hline
         0.03 & $32\times32\times3$ & 0.001 & 128 & 87.27 & 89.50\\ 
         \hline
         0.03 & $72\times72\times3$ & 0.001 & 128 & 88.11 & 89.00\\ 
         \hline
         0.03 & $128\times128\times3$  & 0.001 & 128 & 91.67 & 92.25\\ 
         \hline
         0.03 & $256\times256\times3$  & 0.001 & 128 & 86.43 & 87.75\\ 
         \hline
         0.03 & $128\times128\times3$  & 0.001 & 128 & 88.33 & 89.25\\ 
         \hline
         0.01 & $128\times128\times3$  & 0.001 & 64 & 96.05 & 97.50\\ 
         \hline
         0.01 & $128\times128\times3$ & 0.001 & 32 & 98.62 & 99.25\\ 
         \hline
         0.01 & $128\times128\times3$  & 0.001 & 16 & 94.71 & 95.75\\ 
         \hline
         0.01 & $128\times128\times3$  & 0.001 & 32 & 98.62 & 98.50\\ 
         \hline

    \end{tabular}
    }
    \caption{Manual tuning for Compact Convolutional Transformer}
    \label{tab:baseline2}
\end{table}

Not every hyperparameter shifting obtains better results. Initially, in Table 5, The input size of CCT is $32 \times 32 \times $3, then increases to $72 \times 72 \times $3 and finally to $128 \times 128 \times $3, which gives a value of 91.67\%, thus performing better with the increment of input size. However, after increasing the input size to $256 \times 256 \times $3 the result drops from 91.67\% to 86.43\%. As a result, $128 \times 128 \times $3 of input size demonstrates the optimal result for its parameter. Not only for input size, but the increase in batch size also gives lower results; thus, decreasing it to $32 \times 32 \times $3 shows the most optimal result, as again, keeping the batch size at $16 \times 16 \times $3 shows 3.91\% less accuracy. Thus, keeping it to $32 \times 32 \times $3 is the most optimal tuning possible. 

\begin{table}[ht]
    \centering
    \resizebox{\textwidth}{!}{
    \begin{tabular}{|c|c|c|c|c|}
         \hline
         \textbf{Models} & \textbf{Batch Size} & \textbf{Dropout Rate} & \textbf{Weight Decay} & \textbf{Input Size}\\
         \hline
         ViT & 256 & 0.05 & 0.0001  & $128\times128\times3$ \\
         \hline
         CCT & 32 & 0.1 & 0.05 & $128\times128\times3$ \\
         \hline
         Swin Transformer & 32 & 0.03 & 0.0001 & $72\times72\times3$ \\
         \hline
         EANet & 32 & 0.02 & 0.0001 & $48\times48\times3$ \\
         \hline
         MobileNet & 128 & 0.4 & 0.0001 & $150\times150\times3$ \\
         \hline
         InceptionV3 & 128 & 0.4 & 0.0001 & $150\times150\times3$ \\
         \hline
         EfficientNetB0 & 128 & 0.4 & 0.0001 & $150\times150\times3$ \\
         \hline
         DCECNN & 128 & 0.4 & 0.0001 & $150\times150\times3$ \\
         \hline
    \end{tabular}
    }
    \caption{Tuned hyperparameter values to get optimal results}
    \label{tab:baseline}
\end{table}


\section{Experimental Evaluation}
\label{Experimental_Setup}

\subsection{Experimental Setup}

The necessary libraries that we utilize are Numpy, Keras, Tensorflow, and Matplotlib. The central processing unit (CPU) that we use for model training and dataset collection is an Intel Core i7 12th generation, and the GPU is an NVIDIA GeForce RTX 3070ti with a memory of 12GB and RAM of 64 GB. We run our implementations using Python 3.11.3 on Jupyterlab 3.5.3 as our interactive development environment and harness the capabilities of the TensorFlow 2.6.0 library. To carry out training and testing with our implemented models on our proposed dataset, the dataset is split into train, validation, and test. For training, the proportion is 70\%, for testing it is 20\% and for validation it is 10\%.

\subsection{Experimental Results}

After implementing our architectures on our dataset, we evaluate the performance based on five performance metrics, which are accuracy, precision, recall, F1-score, and Matthews correlation coefficient (MCC), the formulae for which are given below:

\begin{equation}
Accuracy = \frac{TN+TP}{TP+FP+TN+FN}
\end{equation}

\begin{equation}
Precision = \frac{TP}{FP+TP}
\end{equation}

\begin{equation}
Recall = \frac{TP}{FN+TP}
\end{equation}

\begin{equation}
F1 = \frac{2*Precision*Recall}{Recall+Precision}
\end{equation}

\begin{equation}
MCC = \frac{TP \times TN - FP \times FN}{\sqrt{(TP + FP)(TP + FN)(TN + FP)(TN + FN)}}
\end{equation}

The accuracy metric gives us the idea of the many true predictions our implemented model provides out of all the predictions it carries out, which is the sum of true positive (TP) and true negative (TN) divided by the sum of the true positive and true negative, and the false positive (FP) and false negative (FN). For precision, the approach is similar, except it focuses on the positives only, showing how many true positive predictions are made out of both the true and false positives. Recall is also quite similar but focuses on how many predictions are correctly predicted as true by the model. The F1-score is an evaluation metric that combines both precision and recall and is reliable in terms of a class-balanced dataset like the one we propose in our work. We also calculate the Matthews correlation coefficient (MCC) from our obtained true positive and negative values for each model implementation. A higher MCC value represents how well a model is doing in terms of sensitivity and specificity, which means it takes into true positives, true negatives, false positives, and false negatives to provide a summarization of the overall performance of the model.

Implementing the four different transformer-based models, three different neural network-based models, and the DCECNN model consisting of the three CNN models, that too with different input sizes and parameters, we obtain the results as shown in Table \ref{tab:perf metric}.

\begin{table*}[ht]
    \centering
    \resizebox{\textwidth}{!}{
    \begin{tabular}{|c|c|c|c|c|c|}
    \hline
    \textbf{Models} & \textbf{Accuracy} & \textbf{Precision} & \textbf{Recall} & \textbf{F1-score} & \textbf{MCC} \\
    
    \hline
    ViT & 88.66\% & 90.25\% & 89.00\% & 88.00\% &   85.54\% \\
    \hline

    Swin Transformer & 84.74\% & 85.25\% & 85.00\% & 84.75\% &  80.10\%  \\
    \hline

    EANet & 66.56\% & 78.00\% & 66.25\% & 65.00\% &  59.85\%  \\
    \hline

    MobileNet & 70.20\% & 80.75\% & 69.75\% & 64.50\%  &  62.57\% \\
    \hline

    InceptionV3 & 51.83\% & 64.75\% & 58.50\% & 47.75\%  &  42.30\% \\
    \hline

    EfficientNetB0 & 96.32\% & 96.25\% & 96.50\% & 96.00\%  &  95.13\% \\
    \hline
    
     DCECNN &  98.78\% &  98.75\% &  99.00\% &  98.75\% &  98.38\% \\
    \hline

      \textbf{CCT} & \textbf{98.62\%} & \textbf{98.50\%} & \textbf{99.00\%} & \textbf{98.75\%} &  \textbf{98.18\%} \\
    \hline
    
    \end{tabular}
    }
    \caption{Macro average values of accuracy, precision, recall, F1-score, and MCC of the four classes obtained from each model that we implement in our proposed dataset.}
    \label{tab:perf metric}
\end{table*}

In-depth numerical results are shown in Table \ref{tab:perf metric} for every model. The evaluation is carried out based on the popular performance metrics: confusion matrix, precision, recall, and F1-score. As shown in Table \ref{tab:perf metric}, we take the macro average of the accuracy, precision, recall, and F1 score obtained for the four classes.

Looking at the results from the transformer-based models, we can observe that the accuracy, precision, recall, and F1-score are all above 80\% for ViT, CCT, and Swin Transformer, therefore, indicating that these three models perform quite well in terms of delivering accurate results. But when we look at the performance scores of EANet, all of the performance metrics are below 80\%, with accuracy, recall, and F1-score being in the range of 65\% to 67\%. Thus, only EANet shows a considerably lower performance out of the transformer-based models, while the other three architectures show remarkable performance according to the metrics. The values for accuracy and all the other performance metrics peak when CCT is implemented, as we can see the values obtained are all above 98\%. The accuracy for ViT and Swin Transformer also passes the 90\% threshold.

The CNN-based architectures, which are MobileNet, InceptionV3, and EfficientNetB0 show varying results in their performance metrics. InceptionV3 shows the lowest performance as the F1-score, recall, precision, and accuracy are respectively 47.75\%, 58.50\%, 64.75\%, and 51.83\%. MobileNet shows moderate performance with the lowest being the F1-score, which is 64.50\%, and the highest being the precision, which is 80.75\%. EfficientNetB0 gives a remarkable performance with all the performance metrics being over 96\%. For the CNN-based architectures, the precision is the highest out of the other metrics for each model and the F1-score is the lowest. In the DCECNN model, all the metrics show a value of about 99\%, thus showing the best performance, even though not all the models used in the DCECNN model performed well individually. In terms of MCC, we obtain a value higher than 90\% for CCT, DCECNN, and EfficientNetB0, thus indicating a high overall performance compared to the rest of the architectures.

\begin{figure*}[!ht]
\centering
\begin{subfigure}{0.5\textwidth}
\centering
\includegraphics[width = \textwidth, height=\textheight, keepaspectratio]{"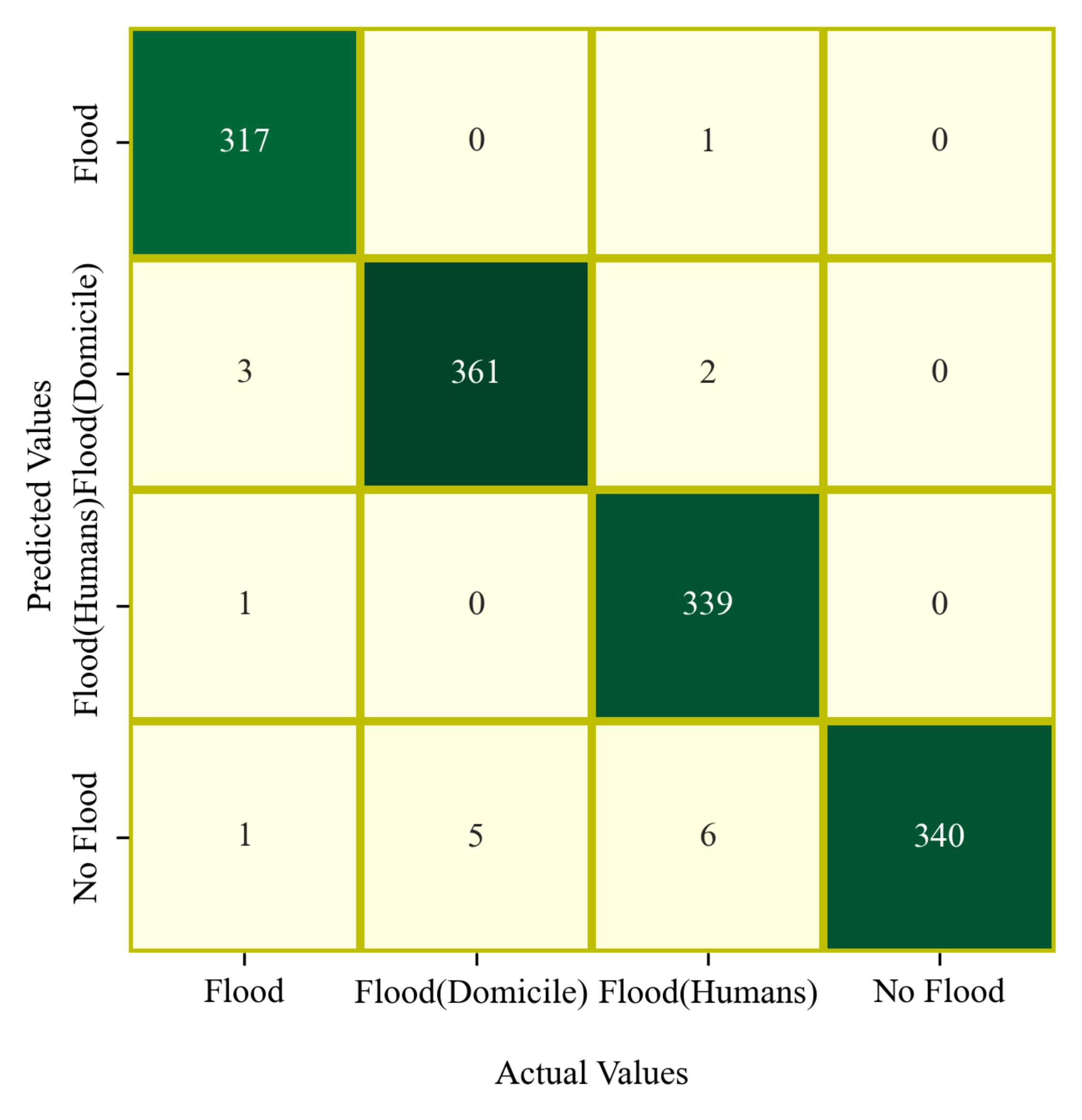"}
\caption{CCT Confusion Matrix}
\label{fig:cm cct}
\end{subfigure}%
\begin{subfigure}{0.5\textwidth}
\centering
\includegraphics[width = \textwidth, height=\textheight, keepaspectratio]{"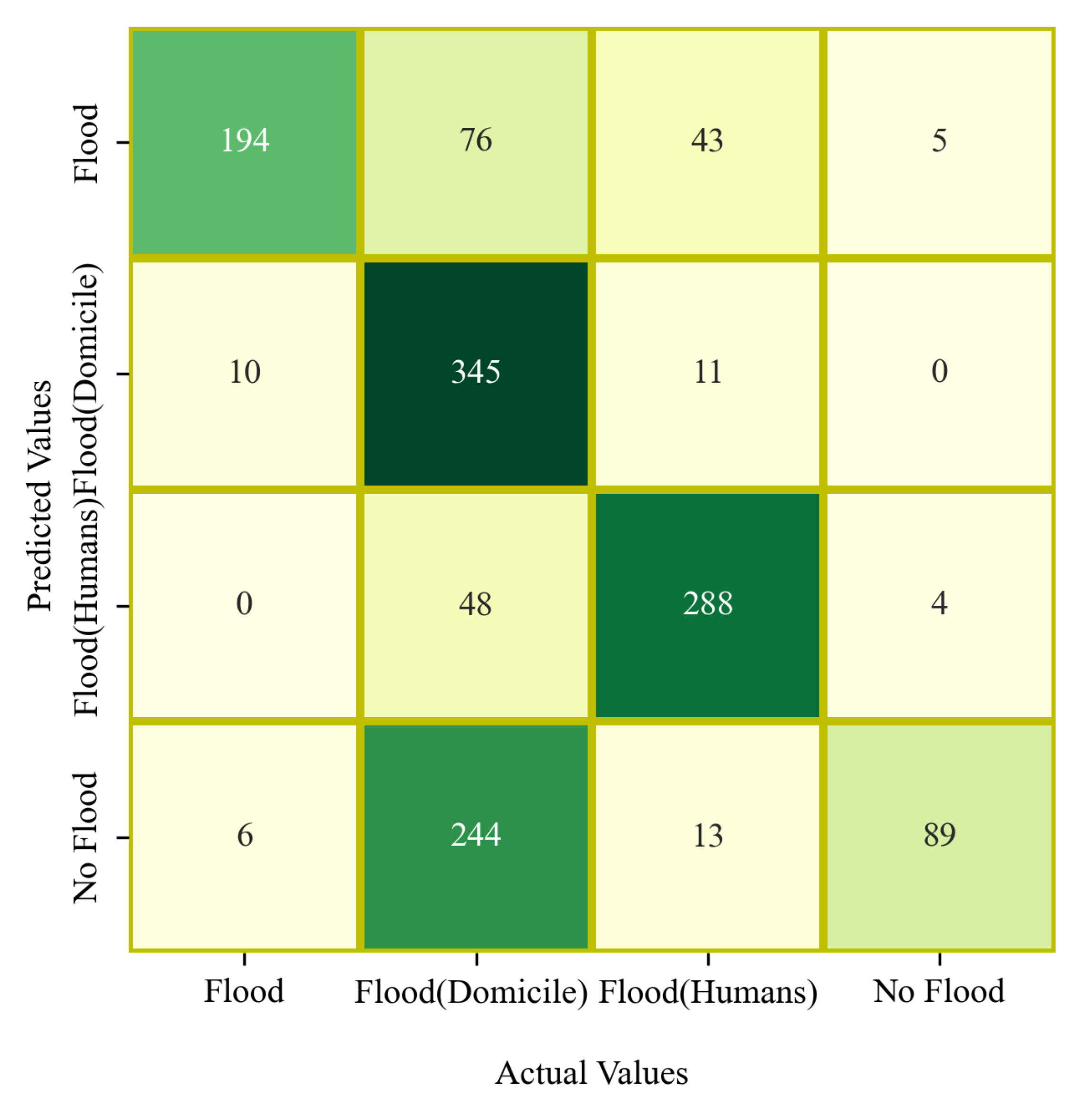"}
\caption{EANet Confusion Matrix}
\label{fig: cm EANet}
\end{subfigure}%
\hfill
\caption{Confusion matrix of CCT and EANet}
\label{fig: cm cct and eanet}
\end{figure*}

The confusion matrix provides a borderline understanding of what or how accurate the classification performance has been. The CCT model reflects the highest accuracy and thus has the most prominent classification results - giving sufficient true positive values. The strongest classification is found in the `flood with domicile' category, a value of 361, followed by `no flood', `flood with humans', and `flood detection' with values of 340, 339, and 317 respectively. The confusion matrix obtained for CCT is shown in Figure \ref{fig:cm cct}. Figure \ref{fig: cm EANet} shows the EANet confusion matrix produces a similar identification outcome, having 345 as the value for the `flood with domicile'. However, the overall matrix is not promising as there is a big conflict in classifying scenarios where there are no floods and with ones containing domiciles, with a score of 244.

\begin{figure*}[!ht]
\centering
\begin{subfigure}{0.45\textwidth}
\centering
\includegraphics[width = \textwidth, height=\textheight, keepaspectratio]{"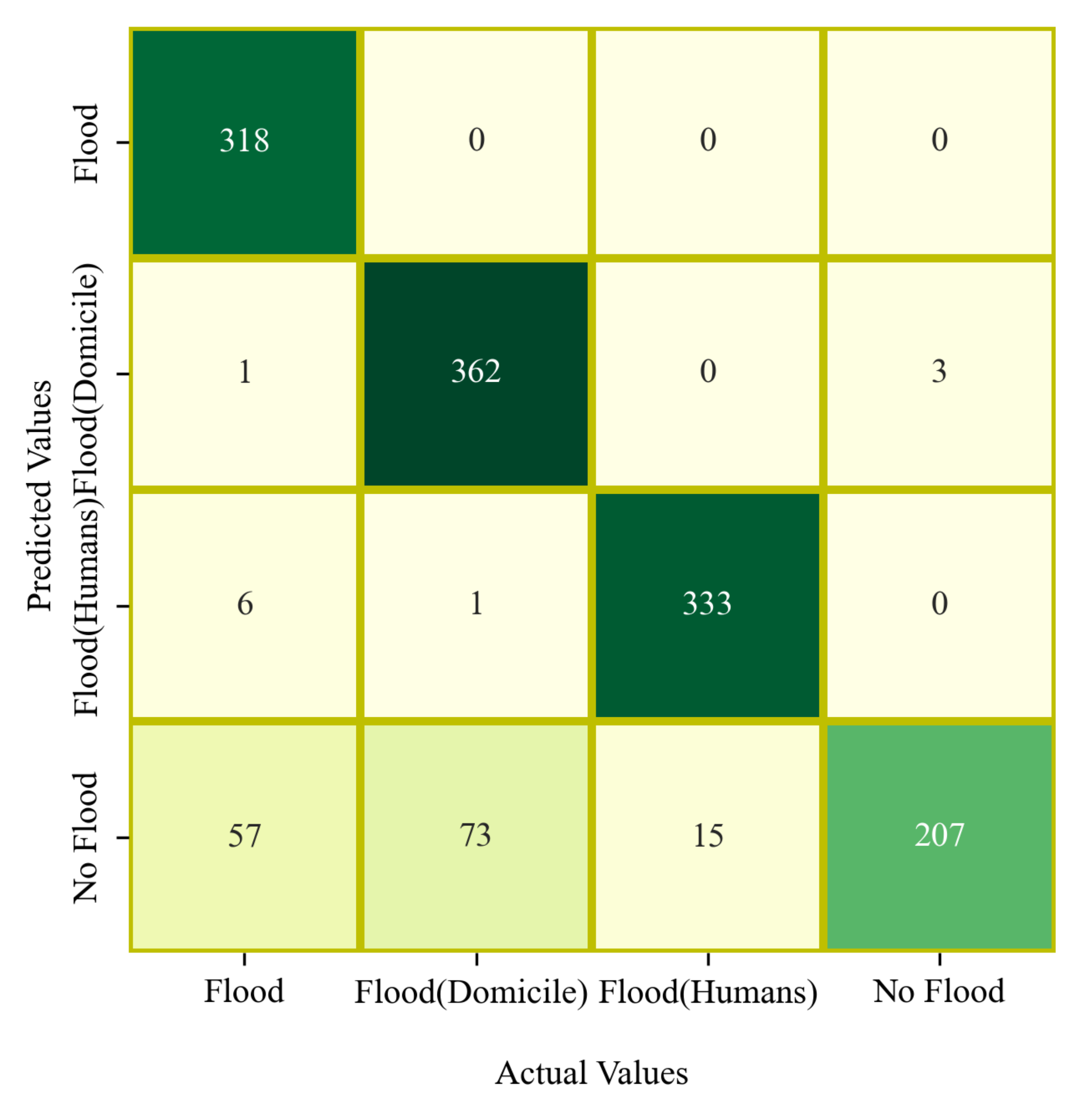"}
\caption{ViT Confusion Matrix}
\label{fig:cm vit}
\end{subfigure}%
\hfill
\begin{subfigure}{0.45\textwidth}
\centering
\includegraphics[width = \textwidth, height=\textheight, keepaspectratio]{"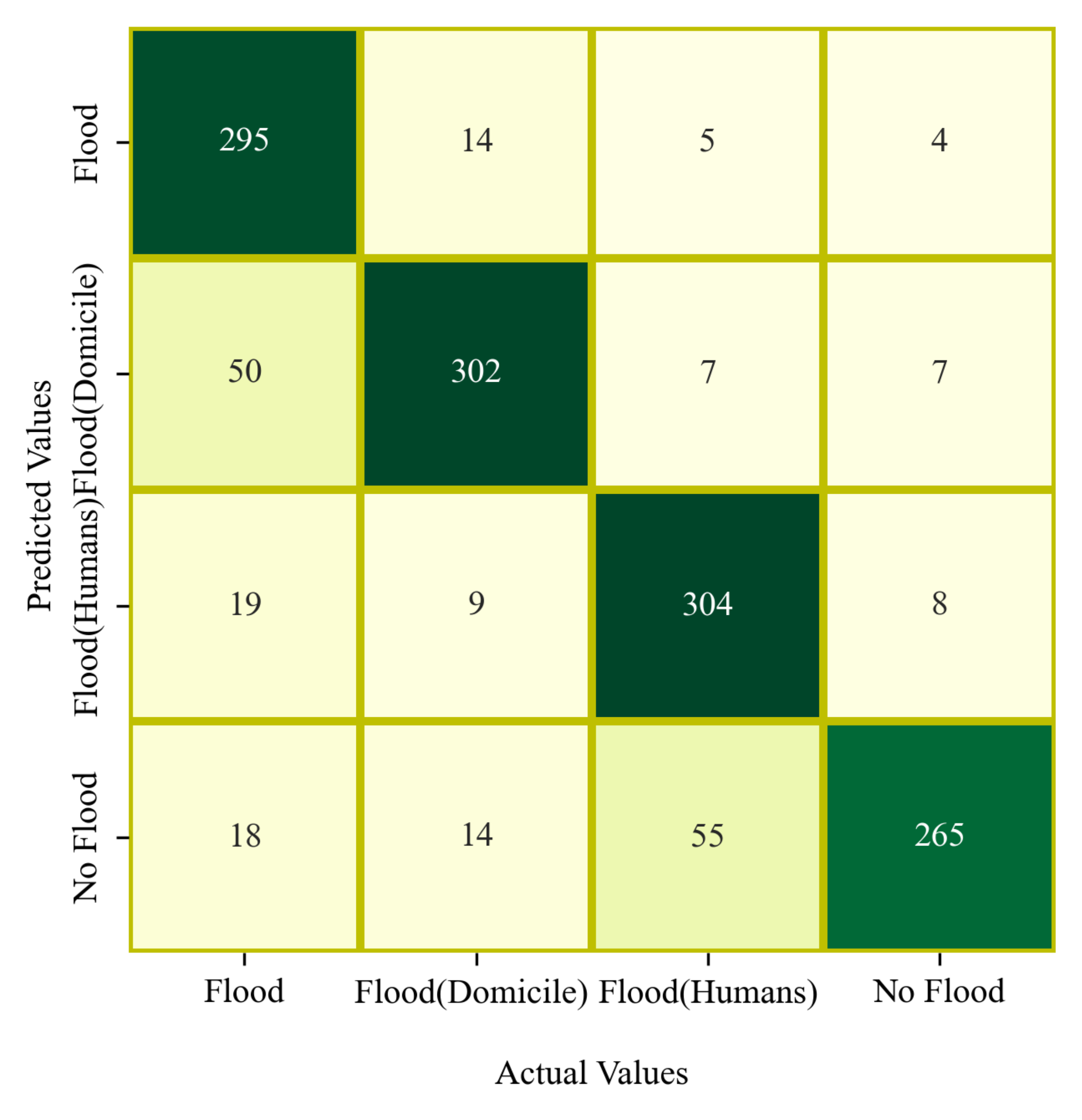"}
\caption{Swin Confusion Matrix}
\label{fig: cm swin}
\end{subfigure}%
\hfill
\begin{subfigure}{0.45\textwidth}
\centering
\includegraphics[width = \textwidth, height=\textheight, keepaspectratio]{"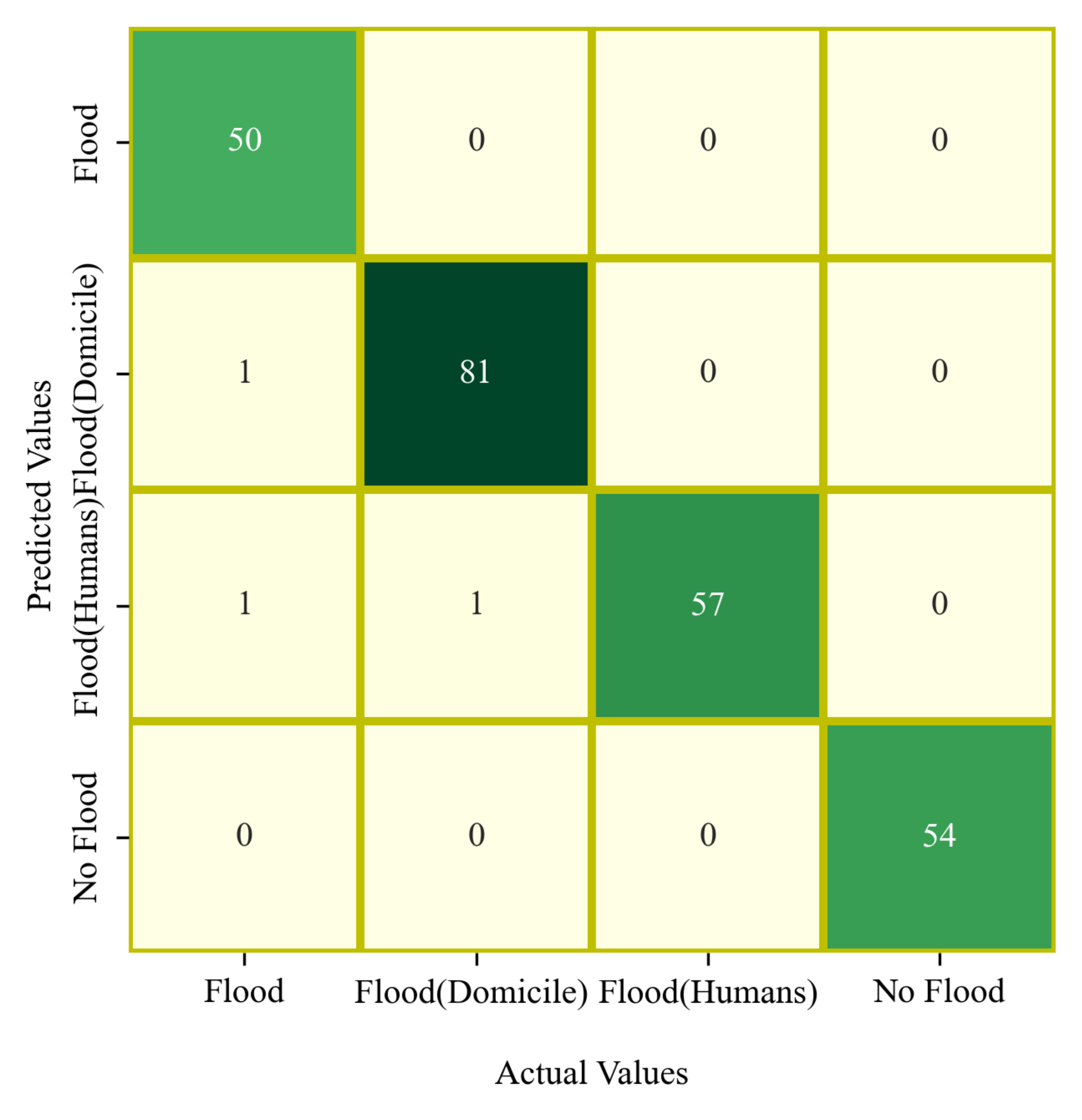"}
\caption{DCECNN Confusion Matrix}
\label{fig: cm ensemble}
\end{subfigure}%
\hfill
\caption{Confusion matrix of ViT, Swin, and EANet}
\label{fig: cm vit-swin-ensemble}
\end{figure*}

Figure \ref{fig:cm vit} shows the ViT model Confusion Matrix which excels in the context of the highest value among every other model - with 362 in flooded domiciles. The obtained values for flood and humans trapped within are good as well, 318 and 333 for each. Some misclassification happens in determining the absence of flood for ViT, as it shows a value of 207, which is a lot less than the other categories. The confusion matrix obtained for ViT is shown in Figure \ref{fig: cm vit-swin-ensemble}. Figure \ref{fig: cm swin} shows Swin Transformer's matrix that represents an exceptional finding to where the highest value is 304, from the `flood with humans' category. The Swin Transformer produces decent classifying outcomes, with the strongest classifications in `flood', `flood with domicile', and `flood with humans', which is almost equivalent to the pattern of the prediction strengths in the ViT confusion matrix. The confusion matrix for the DCECNN model that we form with the aid of MobileNet, InceptionV3, and EfficientNetB0 is shown in Figure \ref{fig: cm ensemble}. Here, only the `flood with domicile' category shows a very strong prediction ability while it remains moderate for the rest of the classes.

\begin{figure*}[!ht]
\centering
\begin{subfigure}{0.45\textwidth} 
\centering
\includegraphics[width=\textwidth, keepaspectratio]{"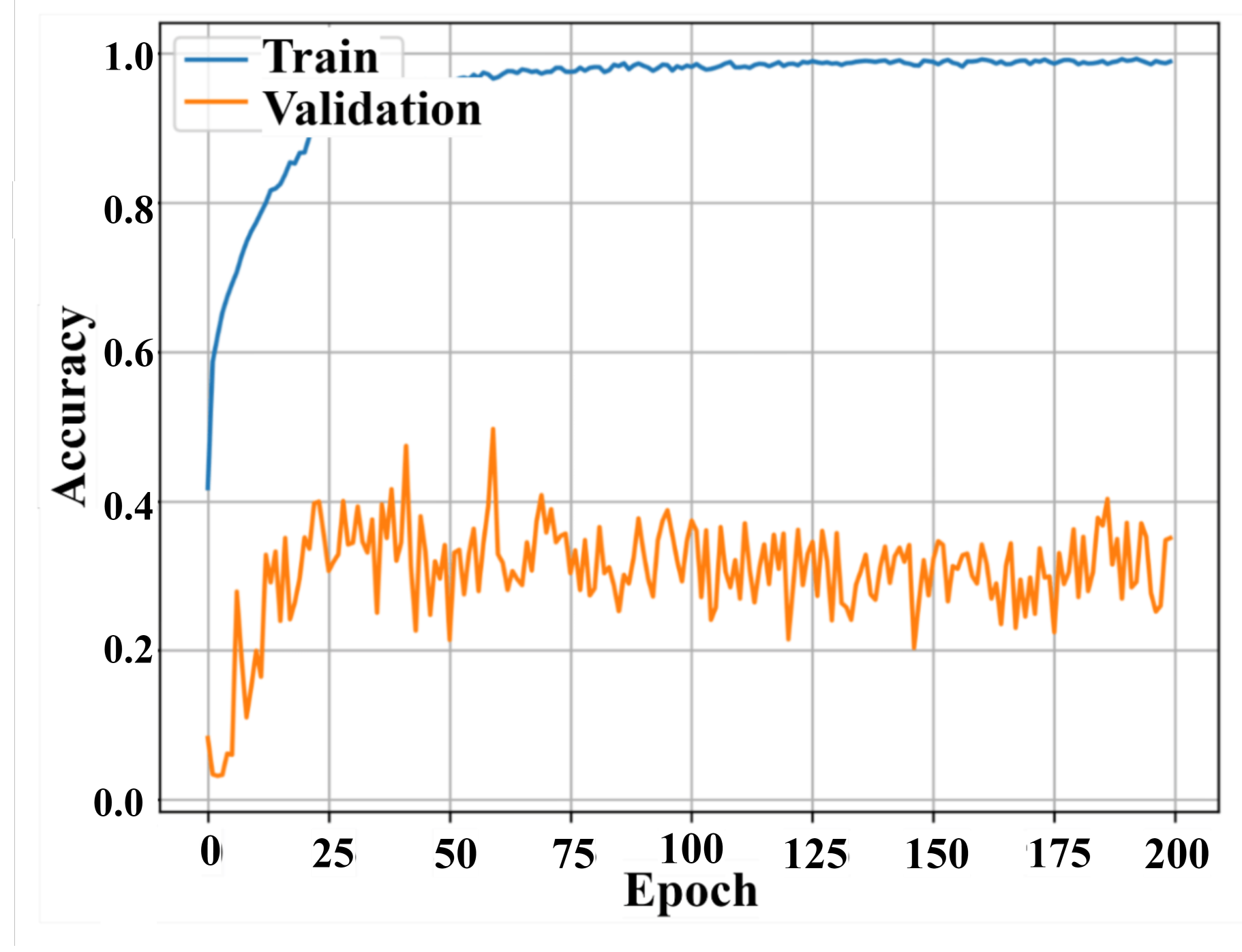"}
\caption{ViT accuracy graph}
\label{fig: a vit}
\end{subfigure}%
\hspace{5mm} 
\begin{subfigure}{0.45\textwidth}  
\centering
\includegraphics[width=\textwidth, keepaspectratio]{"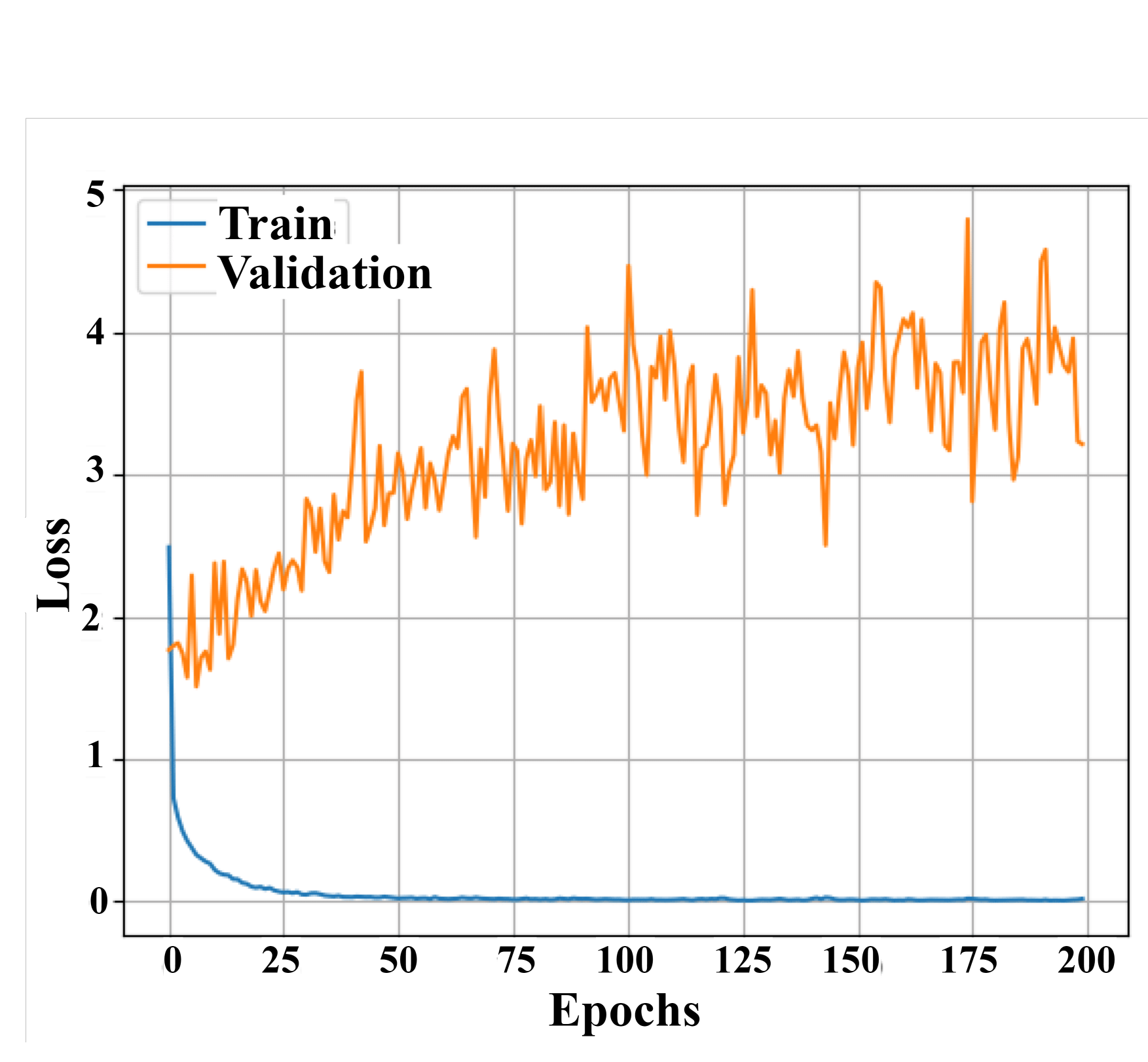"}
\caption{ViT loss graph}
\label{fig: l vit}
\end{subfigure}
\par\bigskip 
\begin{subfigure}{0.45\textwidth}  
\centering
\includegraphics[width=\textwidth, keepaspectratio]{"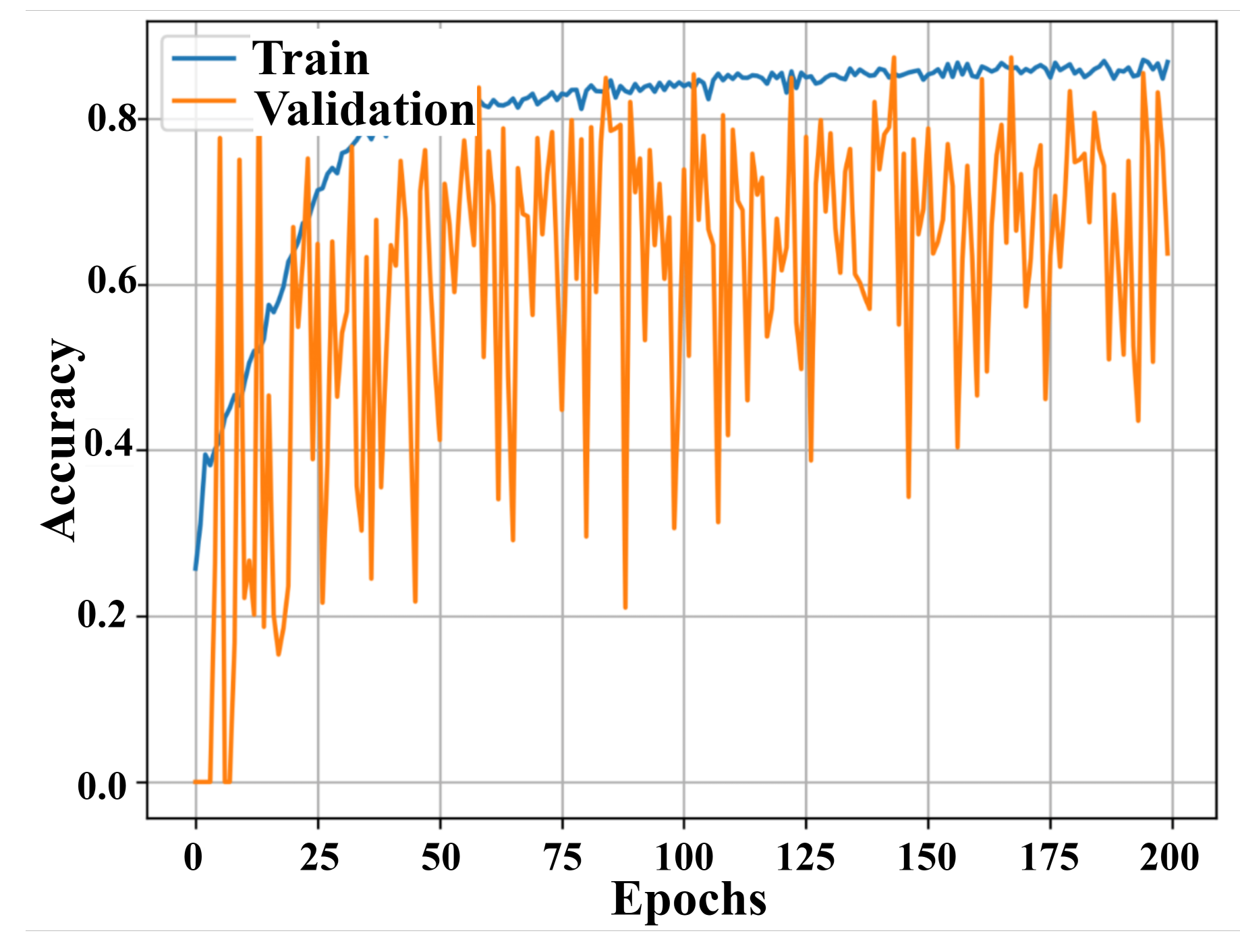"}
\caption{Swin accuracy graph}
\label{fig: a swin}
\end{subfigure}%
\hspace{5mm} 
\begin{subfigure}{0.45\textwidth} 
\centering
\includegraphics[width=\textwidth, keepaspectratio]{"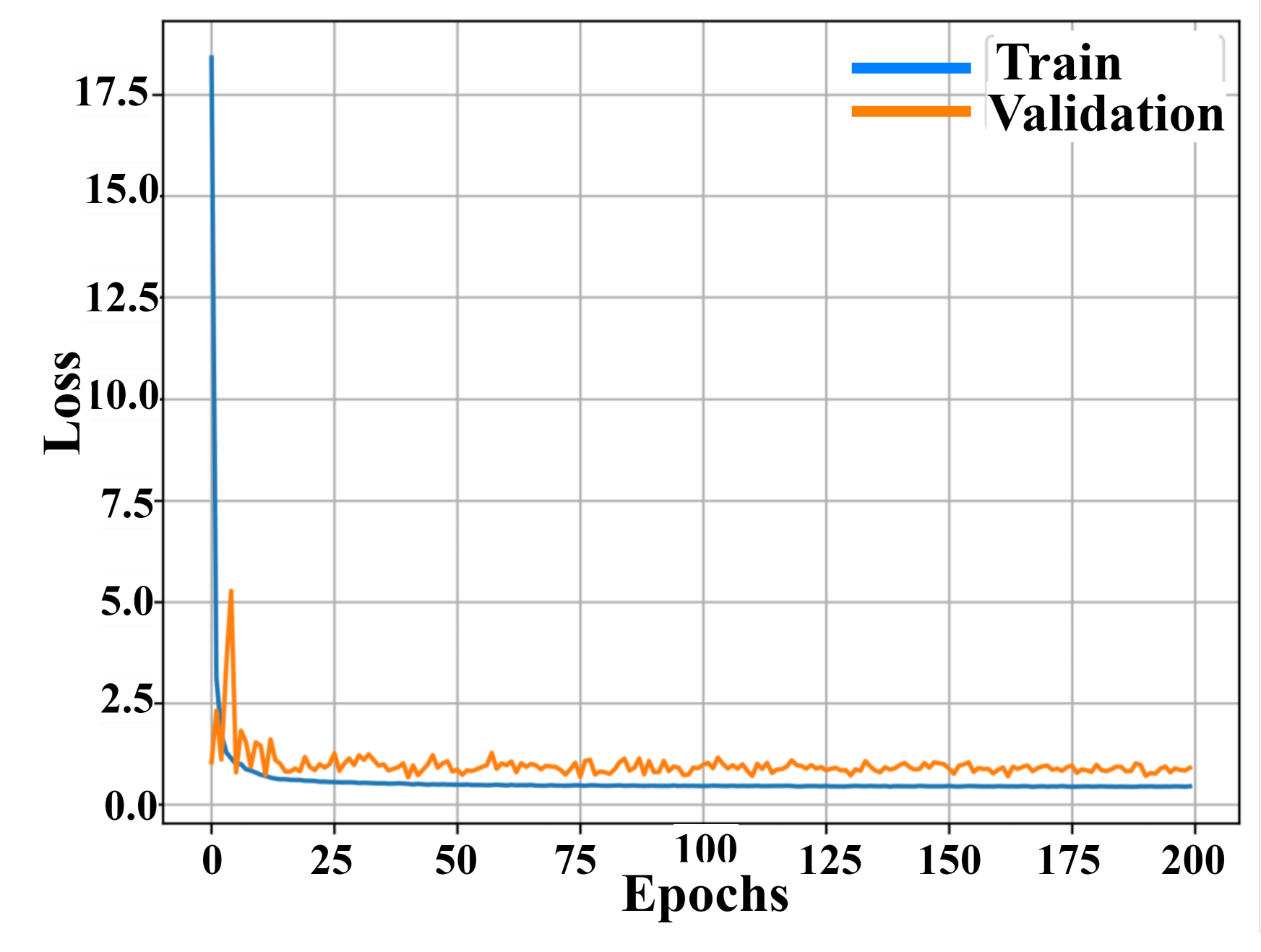"}
\caption{Swin loss graph}
\label{fig: l swin}
\end{subfigure}
\caption{Accuracy and loss graph of ViT and Swin Transformer}
\label{fig: al vit-swin}
\end{figure*}

For each of the models, we have generated training and validation loss graphs which show the train and validation losses that took place over the number of epochs. The graphical representation of training and validation loss gives us an overall picture of the entire learning performance that takes place over the epochs. For example, Figure \ref{fig: l vit} shows the training and validation loss that occurred for Vision Transformers (ViT). In this graph, we can see that the validation loss gets higher over the epochs. This shows that as the epochs go by, the model tends to overfit. The decreasing training loss shows that it performs very well on training data but the opposite for validation data over time. When we look into the train and validation accuracy graph in Figure \ref{fig: a vit}, we can observe a steep rise in the training accuracy over the first few epochs and then it maintains a high accuracy throughout the rest of the epochs. When we look into the validation accuracy curve, a similar pattern but with more fluctuations can be seen, but the benchmark for the highest accuracy reached is almost half of the training accuracy. Since we can see the training accuracy curve being much higher than validation accuracy, we can conclude that overfitting took place. Because, the model performed well on training data but failed to generalize on new data, which is proven by its performance on the validation set.

Figure \ref{fig: l swin} shows the validation and loss graph for the Swin Transformer. Here, due to the decreasing slope for both curves, we can concur that it performs quite well for both training and validation data. The closeness of both the training and validation curve shows the model being neither too overfit nor underfit but close to being optimal fit. When it comes to the accuracy graph of the Swin Transformer in Figure \ref{fig: a swin}, the training and validation curves are much closer compared to that of the other models, for which we can also reach a similar conclusion even though the validation accuracy curve shows fluctuations.

\begin{figure*}[!ht]
\centering
\begin{subfigure}{0.45\textwidth}
\centering
\includegraphics[width=\textwidth, keepaspectratio]{"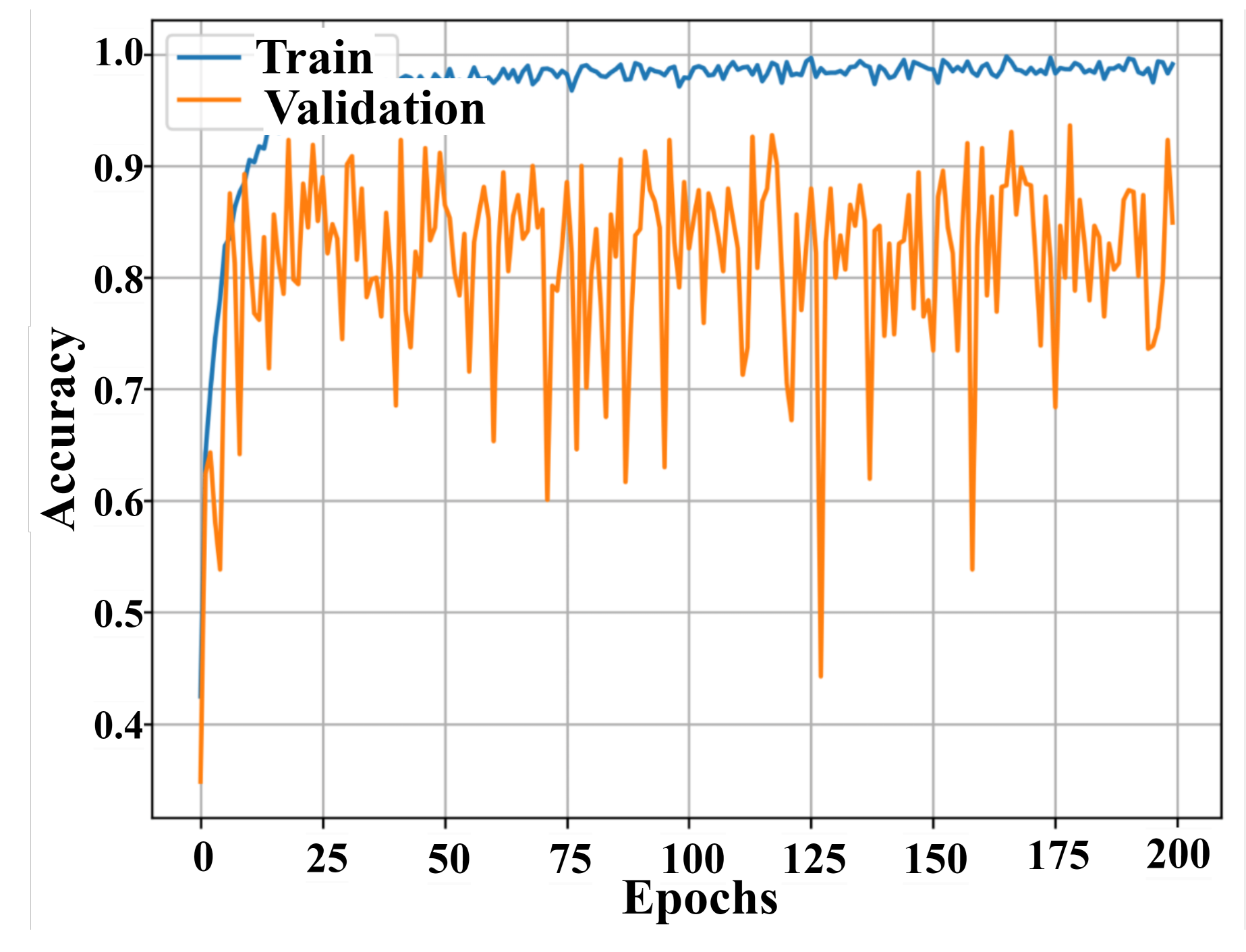"}
\caption{CCT accuracy graph}
\label{fig: a cct}
\end{subfigure}
\hspace{5mm} 
\begin{subfigure}{0.45\textwidth}
\centering
\includegraphics[width=\textwidth, keepaspectratio]{"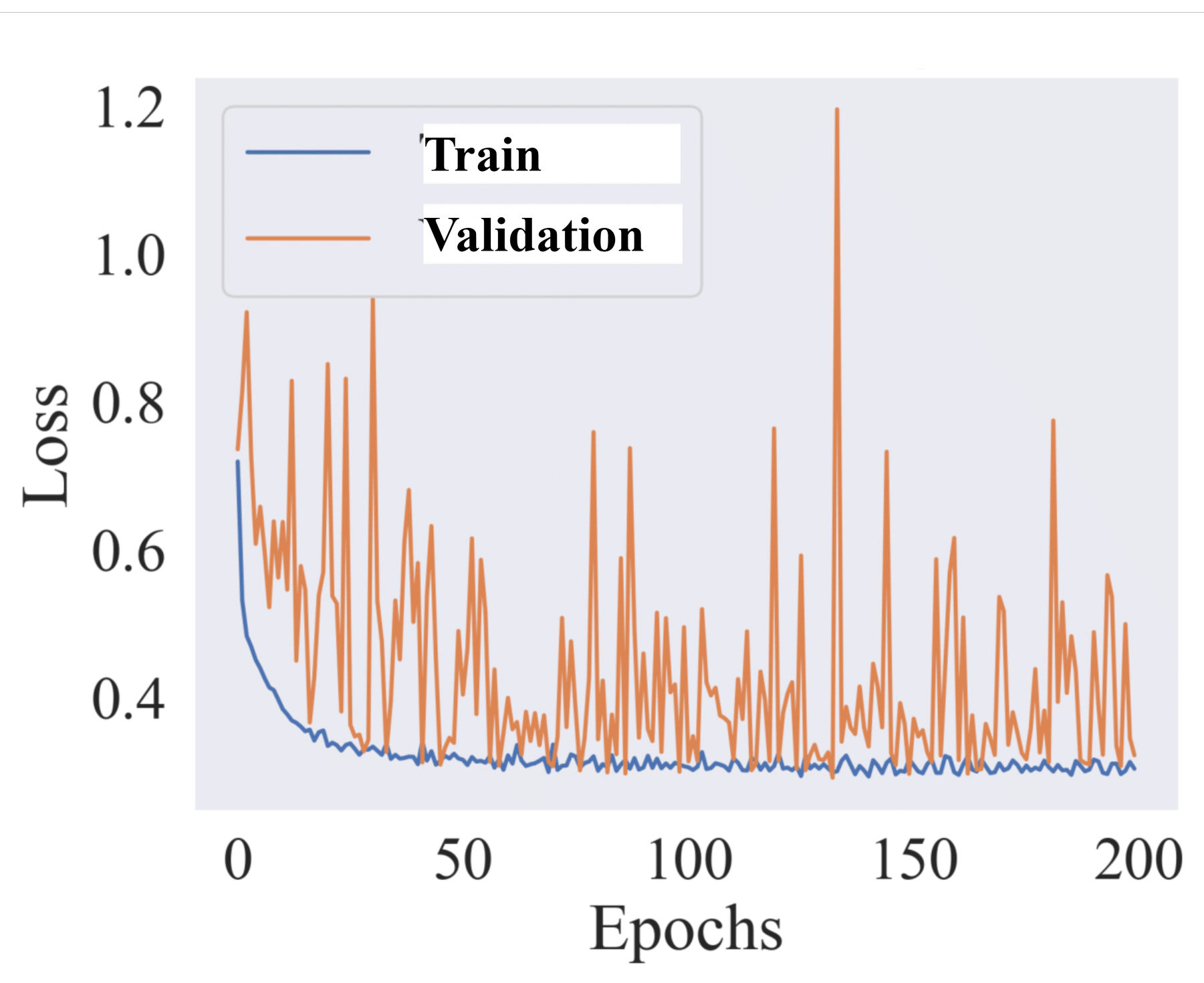"}
\caption{CCT loss graph}
\label{fig: l cct}
\end{subfigure}
\par\bigskip 
\begin{subfigure}{0.45\textwidth}
\centering
\includegraphics[width=\textwidth, keepaspectratio]{"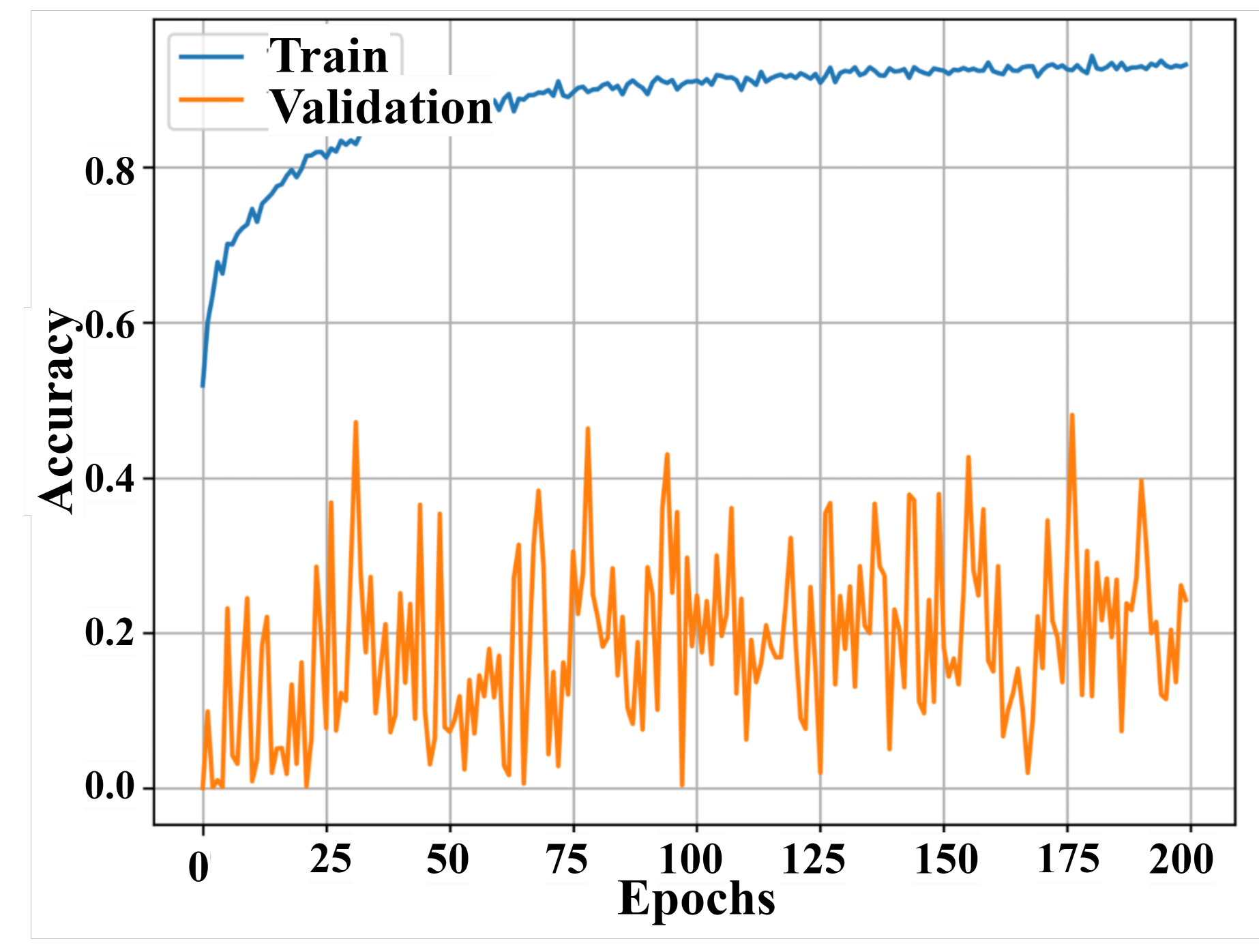"}
\caption{EANet accuracy graph}
\label{fig: a eanet}
\end{subfigure}
\hspace{5mm}
\begin{subfigure}{0.45\textwidth}
\centering
\includegraphics[width=\textwidth, keepaspectratio]{"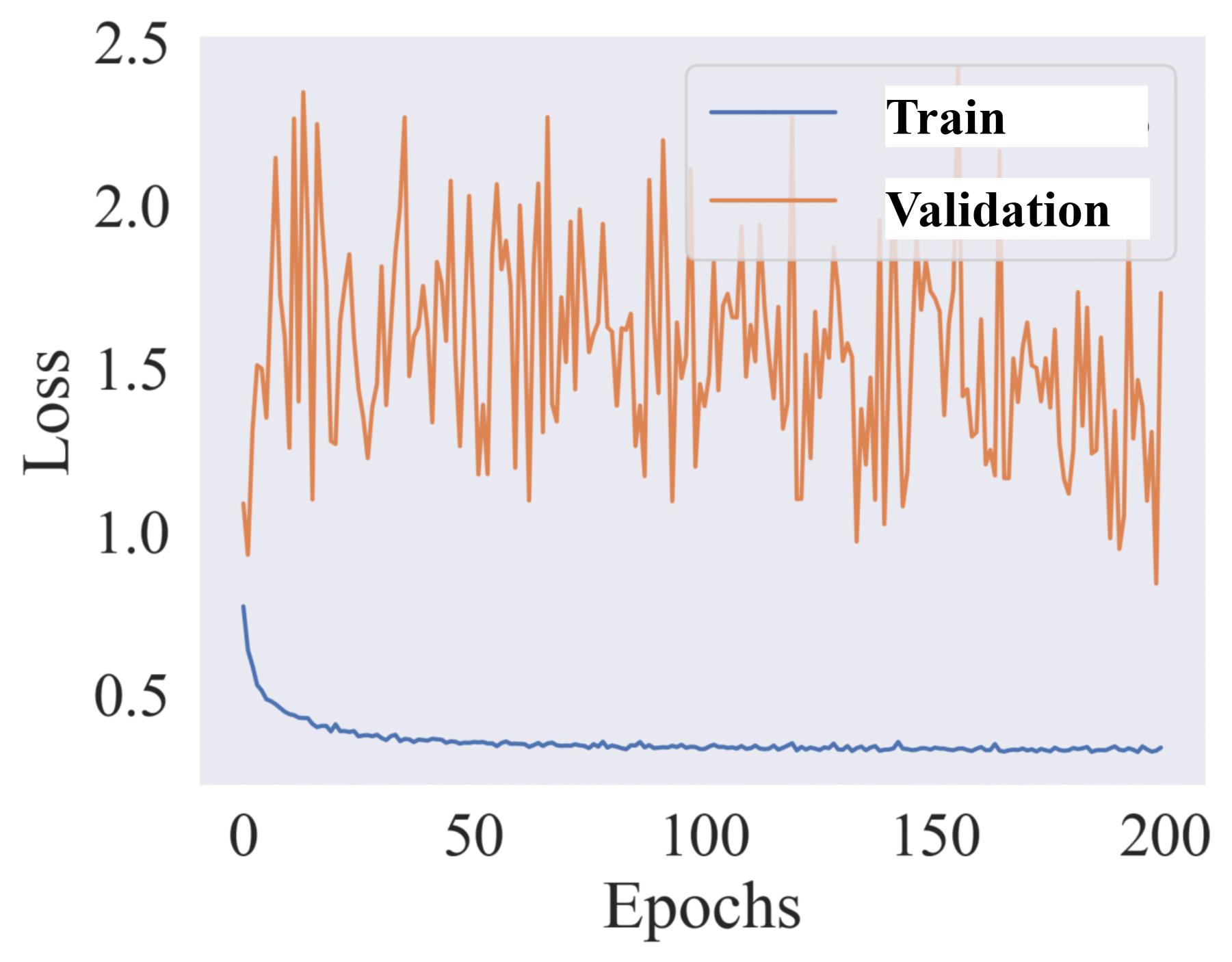"}
\caption{EANet loss graph}
\label{fig: l eanet}
\end{subfigure}
\caption{Accuracy and loss graph of CCT and EANet}
\label{fig: al cct-eanet}
\end{figure*}

Figure \ref{fig: l cct} shows the training and validation graph of Compact Convolutional Transformers (CCT), which shows a fluctuating validation loss curve and training loss curve with a downward slope. This shows that it performs well for training data and validation data. While the performance fluctuated for validation data, it remained close to the training loss curve for which it did not overfit much. The accuracy graph for CCT in Figure \ref{fig: a cct} shows the low gap between the training and validation curve, with the training curve being slightly higher after the first few epochs, thus showing the presence of little overfitting.

Finally, Figure \ref{fig: l eanet} also shows the graph for EANet, where we can also see a fluctuating validation curve and a training loss curve with a downward slope. The train loss curve thus represents good performance on training data and since the fluctuating validation curve is far above the training curve, we can concur that overfitting occurred for this model. A huge gap between the training and validation curve can also be observed in the accuracy graph in Figure \ref{fig: a eanet}, where the training curve is much higher than the validation curve, thus indicating the occurrence of overfitting.

\begin{figure*}[ht]
\centering
\begin{subfigure}{0.45\textwidth}
\centering
\includegraphics[width = \textwidth, height=\textheight, keepaspectratio]{"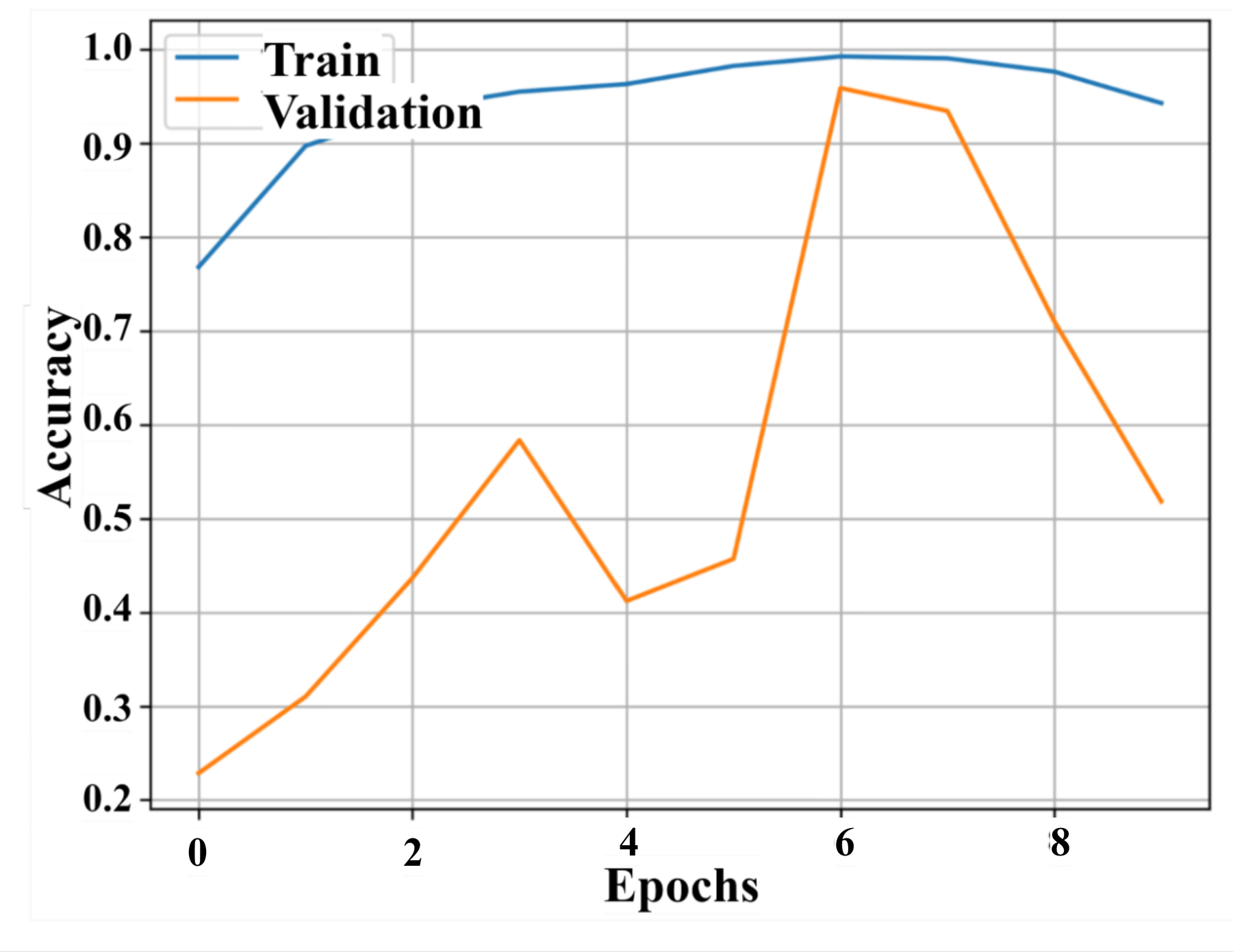"}
\caption{DCECNN  accuracy graph}
\label{fig: a ensemble}
\end{subfigure}%
\begin{subfigure}{0.45\textwidth}
\centering
\includegraphics[width = \textwidth, height=\textheight, keepaspectratio]{"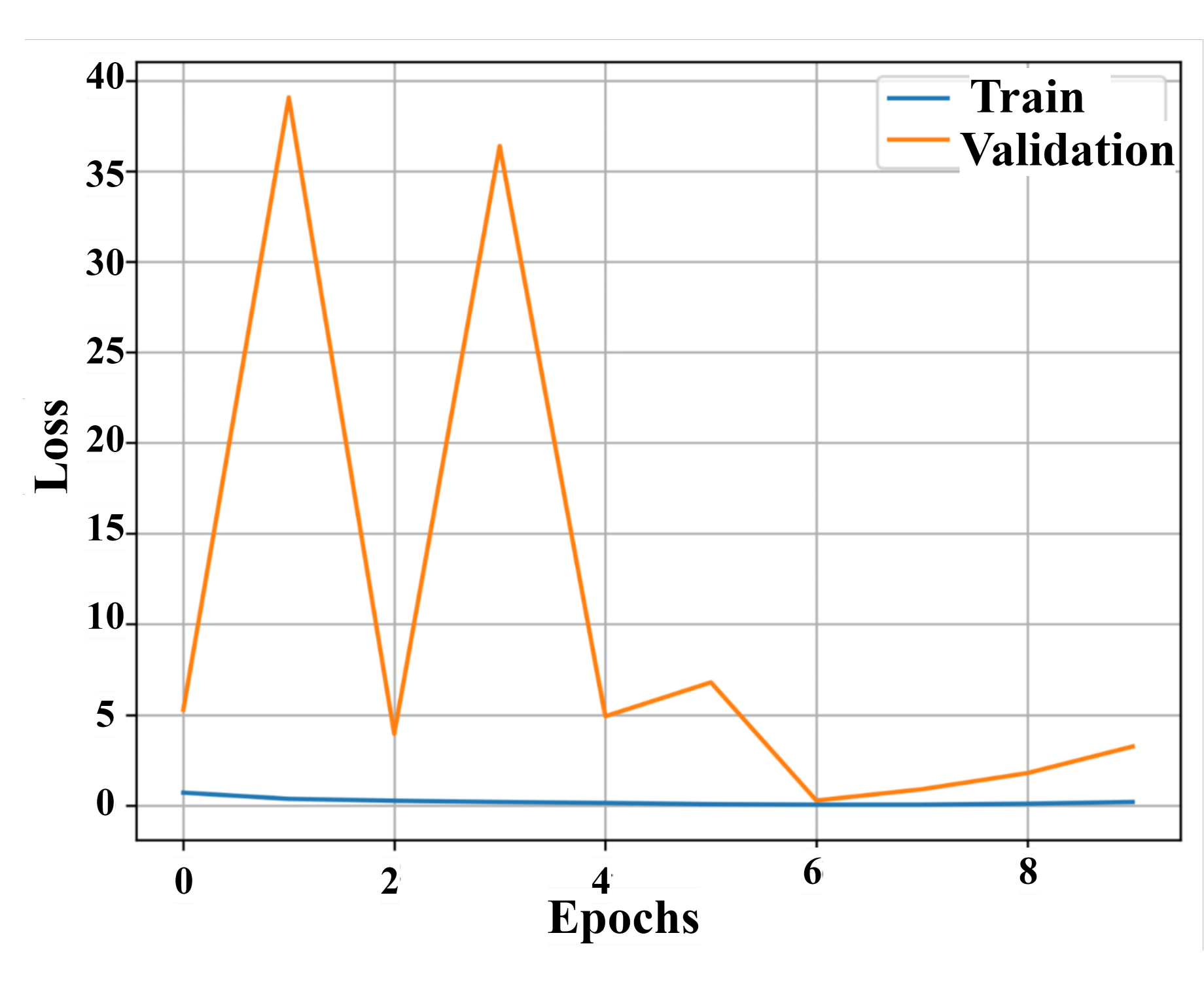"}
\caption{DCECNN loss graph}
\label{fig: l ensemble}
\end{subfigure}%
\caption{Accuracy and loss graph of the DCECNN}
\label{fig: al ensemble}
\end{figure*}

In the training and validation loss graph of the DCECNN model in Figure \ref{fig: l ensemble}, we can observe the validation loss being very high at the beginning. In contrast, the training loss almost remains the same throughout the epochs. The validation loss curve fluctuates for each epoch but the loss decreases towards the end and the curve gets near the training loss curve, thus lowering the overfitting over time. The same pattern can also be observed when looking at the accuracy graph of the DCECNN model in Figure \ref{fig: a ensemble}, as we can observe the higher gap between the training and validation curves at first, with the training curve above the validation curve. But towards the last epochs, the validation curve gets closer to the training curve.

We have provided explicit details on the flops,  training time per epoch and inference time for each transformer-based model in Table \ref{tab: time and flops}, as a keen understanding of computing efficiency is crucial for practical deployment and scalability.

\begin{table*}[ht]
    \centering
    \resizebox{\textwidth}{!}{
    \begin{tabular}{|c|c|c|c|}
    \hline
    \textbf{Models} & \textbf{FLOPs (G)} & \textbf{Training time (s)} & \textbf{Inference time (s)} \\
    
    \hline
    CCT & 0.896 & 13.069  & 1.023  \\
    \hline
    
    ViT & 1.03 & 2.087  & 1.017  \\
    \hline

    Swin Transformer & 0.091 & 9.046  & 1.014  \\
    \hline

    EANet & 0.154 & 13.074  & 2.034  \\
    \hline
    
    \end{tabular}
    }
    \caption{FLOPs, training time and inference time obtained from each transformer model we implement in our proposed dataset.}
    \label{tab: time and flops}
\end{table*}

\subsection{Experimenting on the FloodNet Dataset}

\begin{table*}[ht]
    \centering
    \resizebox{\textwidth}{!}{
    \begin{tabular}{|c|c|c|c|c|c|c|c|c|}
    \hline

    \multicolumn{4}{|c|}{\textbf{Proposed Dataset}} &  & \multicolumn{4}{|c|}{\textbf{FloodNet Dataset}} \\
    \hline
    
    \textbf{F1-score} & \textbf{Recall} & \textbf{Precision} & \textbf{Accuracy} & \textbf{Models} & \textbf{Accuracy} & \textbf{Precision} & \textbf{Recall} & \textbf{F1-score} \\
    
    \hline
    82.50\% & 79.50\% & 86.50\% & 88.69\% & ViT & 60.68\% & 63.50\% & 63.00\% & 60.50\%  \\
    \hline
 
    98.50\% & 98.50\% & 99.00\% & 99.24\% & CCT & 88.47\% & 88.00\% & 88.50\% & 88.00\%  \\
    \hline

    83.50\% & 85.00\% & 85.00\% & 84.08\% & Swin Transformer & 78.98\% & 79.50\% & 76.00\% & 77.00\% \\
    \hline

    56.50\% & 43.00\% & 91.50\% & 91.88\% & EANet & 78.31\% & 78.50\% & 76.50\% & 77.00\% \\
    \hline

    44.00\% & 56.50\% & 73.00\% & 49.56\% & MobileNet & 91.94\% & 91.50\% & 92.50\% & 91.50\% \\
    \hline

    61.50\% & 87.50\% & 47.00\% & 47.40\% & InceptionV3 & 79.15\% & 78.50\% & 79.50\% & 78.50\% \\
    \hline

    95.00\% & 97.00\% & 94.00\% & 93.52\% & EfficientNetB0 & 97.63\% & 98.00\% & 97.50\% & 97.50\% \\
    \hline

    99.00\% & 99.50\% & 98.00\% & 98.11\% & DCECNN  & 95.97\% & 95.50\% & 96.00\% & 96.00\% \\
    \hline
    
    \end{tabular}
    }
    \caption{A comparison between the values of the performance metrics that we obtain with our proposed dataset and the FloodNet \cite{9460988} dataset. Here, the precision, recall, and F1-score values for both datasets are the macro average of the `flood' and `no flood' classes only.}
    \label{tab:comparison_floodnet}
\end{table*}

Rahnemoonfar et al. \cite{9460988} propose a dataset that is quite similar in general to our proposed dataset, which consists of aerial imagery of flood-affected areas during the occurrence of Hurricane Harvey in the USA. The paper also demonstrates image classification, segmentation, and visual question and answer. When it comes to classification, the categories that the authors mainly aim to classify are flooded and non-flooded areas. So, the images contained in this dataset are mostly capable of carrying classification for those two categories only. Given how the images in this dataset mostly contain houses that are not immersed in flood but have flood water nearby and also with no sign of human presence, we cannot perform house and human classification with the same approach that we undertake in our proposed dataset. So, we train and test the same models using the same parameters and input size with this dataset and classify taking only the `flood' and `no flood' categories. We also carry out fine-tuning for each model to get the optimum results for this dataset, like how we did when implementing them in our proposed dataset. Table \ref{tab:comparison_floodnet} shows the results of the same performance metrics that we use for our dataset. The values of accuracy, precision, recall, and F1-score are also the macro average of the values obtained for the `flood' and `no flood' classes. So, to provide a fair comparison, instead of taking the macro average of four classes from the implementation in our proposed dataset AFSSA, we take the macro average of the `flood' and `no flood' classes only, since no human and house classification is conducted with the FloodNet dataset.

If we look at the results, we observe that the testing accuracy, precision, recall, and F1-score from the FloodNet dataset for ViT is within the range of 60\% to 64\%, which is quite low compared to the results received from training and testing our proposed dataset with ViT, which ranged from 89\% to 90\%. The same goes for the performance metrics of CCT and Swin Transformer, where the values obtained for FloodNet are lower than those of our proposed dataset. But in the case of EANet, the results are the opposite as FloodNet shows better results. Now, for the CNN-based architectures that we implement independently and then as an ensemble model, the results of the performance metrics are in favor of the FloodNet dataset. When we train and test the MobileNet, InceptionV3, and EfficientNetB0 architecture, with the FloodNet dataset individually, it gives a remarkably higher and better result than the results that we obtain from our proposed dataset. For example, the accuracy for MobileNet jumped from 49.56\% to 91.94\%, for InceptionV3 it jumped from 47.40\% to 79.15\% and for EfficientNetB0 it went from 93.52\% to 97.63\%. Comparing the results of Ensemble for both the datasets, our dataset shows a slightly better result as all the performance metrics show a result of approximately 99\% while for the FloodNet dataset, the values are around approximately 96\%. Table \ref{tab:comparison_floodnet} shows the values of the performance metrics that we obtain from both datasets side-by-side for each implemented model.

The results that we obtain with both the FloodNet dataset and our custom dataset AFSSA give a general overview of which type of architecture works better with our dataset. For the DCECNN, both the datasets give similar results thus showing that they both perform almost similarly with this model. Comparing the performance of the CNN architectures individually, FloodNet has the upper hand as it seems to show better performance. When it comes to the transformer-based architectures, we can see that three out of four of them show better performance with our dataset, which hints that our dataset is likely to suit well for transformer-based architectures designed for computer vision.

\subsection{YOLOv8 Implementation}

\begin{figure}[ht]
\centerline{\includegraphics[scale=0.35]{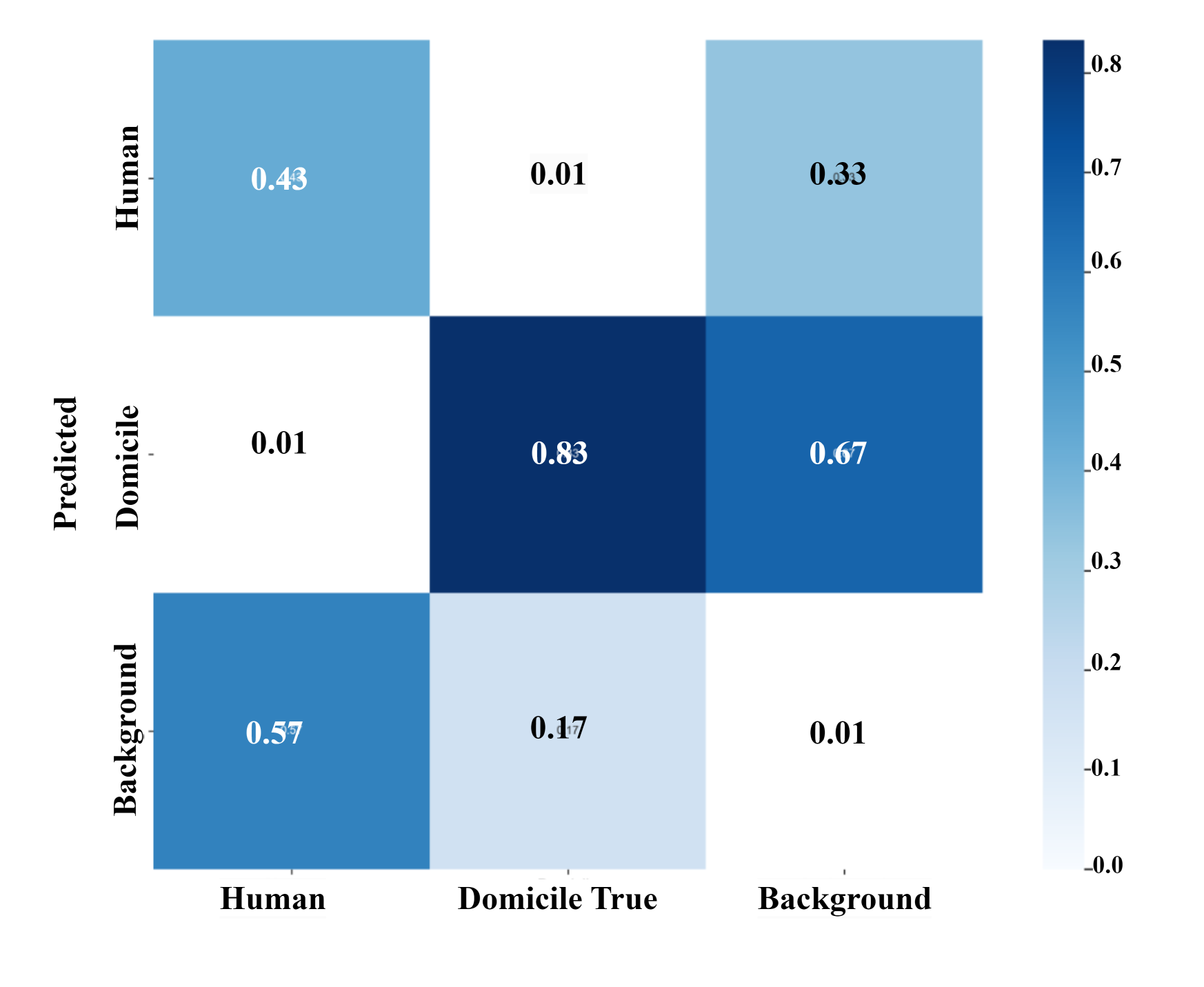}}
\caption{Confusion Matrix obtained from YOLOv8 implementation}
\label{cm detection}
\end{figure}

Firstly, since we are observing the performance of detection, it is more applicable to detect specific objects within a frame, like humans and domiciles, as flood zones are hard to include within the bounding boxes. So, we take the images of the `flood with humans' and `flood with domicile' categories from our dataset and divide them into training, validation, and testing. The model performs detection on the testing images through the use of bounding boxes to identify our desired labels, which are the two categories we are aiming to detect. Hence, during training, the bounding boxes that we label are `Domicile' and `Human'.  Figure \ref{fig: all detections} below shows the prediction results obtained from three testing images. For each prediction, a confidence score is also obtained and displayed. Based on the confidence scores of each prediction of both categories, we can see that it can detect domiciles within these images better than humans, as we can see in Figure \ref{fig: all detections} that not all humans are detected on the boat. The confidence score also fluctuates largely between 1.00 to 0.50, which indicates that the model requires further training data and much higher parameters to carry out a decent level of prediction. Figure \ref{cm detection} shows the confusion matrix obtained from our implementation of YOLOv8, which shows the normalized representation of how many true predictions the model makes for humans and domiciles, and also for the background region where the model doesn't detect any of the objects. As we can see in the confusion matrix, about 83\% of the instances where domiciles are present are correctly detected, which represents the highest number of detections for any of the categories.

\begin{figure*}[ht]
\centering
\begin{subfigure}{0.45\textwidth}
\centering
\includegraphics[width = \textwidth, height=\textheight, keepaspectratio]{"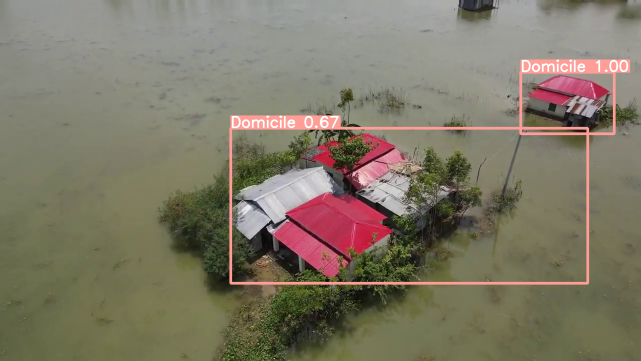"}
\label{fig:detection_2}
\end{subfigure}
\hfill
\begin{subfigure}{0.46\textwidth}
\centering
\includegraphics[width = \textwidth, height=\textheight, keepaspectratio]{"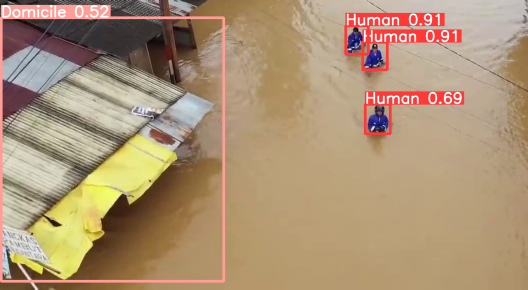"}
\label{fig:detection_3}
\end{subfigure}
\hfill
\begin{subfigure}{0.45\textwidth}
\centering
\includegraphics[width = \textwidth, height=\textheight, keepaspectratio]{"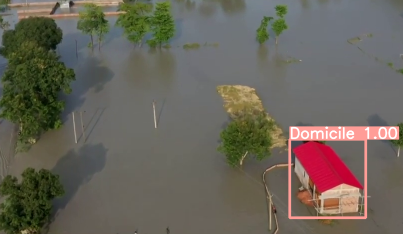"}
\label{fig: detection_5}
\end{subfigure}
\hfill
\begin{subfigure}{0.45\textwidth}
\centering
\includegraphics[width = \textwidth, height=\textheight, keepaspectratio]{"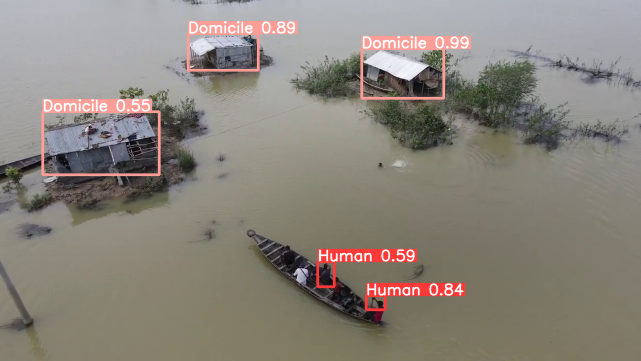"}
\label{fig: detection_1}
\end{subfigure}
\hfill
\caption{Detection results obtained from YOLOv8 implementation}
\label{fig: all detections}
\end{figure*}

\begin{table*}[!ht]
    \centering
    \resizebox{\textwidth}{!}{
    \begin{tabular}{|p{2.5cm}|p{2.8cm}|p{2cm}|p{3cm}|p{4.5cm}|p{5cm}|}
    \hline
    \textbf{Research Study} & \textbf{Prime Architecture} & \textbf{Dataset} & \textbf{Performance} & \textbf{Research Study Limitations} & \textbf{Improvements in Our Study}\\
    
    \hline
    Rahnemoonfar et al. \cite{9460988}, 2020 & InceptionNetv3, ResNet50 and Xception & Custom Dataset & 84.38\%, 93.69\% and 90.62\% Accuracy & The FloodNet dataset this study proposes consists of flood imagery that is specific to a single country and a single flooding event, which is the Hurricane Harvey in the USA.  &  We propose a dataset that consists of imagery from various flooding events taking place in several flood-prone countries specifically in South Asia, which includes Bangladesh, India, Pakistan and Indonesia. \\
    \hline

    Seydi et al. \cite{rs15010192}, 2022 & SVM, Decision Tree, Random Forest, (DNN), LightGBM, XGBoost and CFM & Custom Datset & 94\%  Accuracy & Compares only traditional individual machine learning models by obtaining an accuracy of no more than 94\% & We construct DCECNN, a custom ensemble architecture with three state-of-the-art CNN models, which results in an optimal combination that achieves a notable performance of 98.78\% accuracy and almost 99\% of precision, recall and F1-score. \\
    \hline
    
    Akshya et al. \cite{akshya2019hybrid}, 2019 &  SVM and k-means clustering & Custom Dataset & 92\% Accuracy &  Classifies only two classes (flood-affected areas and non-flood affected areas) &  We introduce four classes, the two extra classes are 'flood with humans' and 'flood with domicile'. \\ 
    \hline

    Jackson et al. \cite{w15050875}, 2023 & Vision Transformer & FloodNet Dataset & 96.931\% accuracy and an F1-score and precision of around 88\% & Only applies Vision Transformer and obtains an accuracy of 96.931\% and an F1-score and precision of 88\%. The ViT models implemented have a large number of parameters (86 million). & We implement four different transformer-based models (ViT, CCT, Swin Transformer, and EANet), among which CCT gives 98.62\% accuracy and precision, recall, and F1-score of almost 99\%, which exceeds the performance of the aforementioned study. The CCT model that we implement consists of a very low number of parameters (0.41 million only).\\
    \hline

    Munawar et al. \cite{munawar2021application}, 2021 & Proposed CNN architecture & Custom Dataset & 91\% Accuracy &  Use of only noise removal and orthorectification when it comes to images preprocessing & We implement CLAHE (Contrast-Limited Adaptive Histogram Equalization) for preprocessing dataset images to enhance the local contrast of an image by dispersing the intensity levels in certain areas. \\
    \hline

    Dong et al. \cite{dong2021uav}, 2021 & YOLOv3-Darknet & N/A & 85.03\% of Mean Average Precision & YOLOv3 is not recent or state of the art & We implement YOLOv8m, a recent and cutting-edge state of the art YOLOv8 model.\\
    \hline
    
    \end{tabular}
    }
    \caption{Comparison table with recent related works}
    \label{tab: recent comparison}
\end{table*}

\section{Discussion}
\label{Discussion}

In this section, we first present a comparison with recent works to contextualize our findings within the existing body of literature. Following this, we delve into a comparative analysis between classification and detection methodologies, highlighting their respective strengths and limitations. We then summarize the key findings of our study, emphasizing the contributions and innovations introduced. Finally, we address the limitations of our research, outlining potential areas for future improvement and exploration.

\subsection{Comparison with Recent Works}

Table \ref{tab: recent comparison} represents recent studies that are based on flood imagery or identifying humans or houses. The table shows the prime architecture each paper has implemented and the level of performance achieved by them. The table also acknowledges whether recent studies introduce a custom dataset for their implementations or outsource the dataset, in that case, we include the name of the dataset they have used. For each study, we showcase the drawbacks of each paper compared to our work and then concurrently, we provide the specific contributions we make in our work that counter those drawbacks.

\subsection{Comparison Between Classification and Detection}

Among our implementations, the transformer-based architectures and the CNN-based architectures are used to carry out image classification and the YOLOv8 model is used to carry out detection of certain objects, which are houses and humans. In terms of performance, The YOLOv8 model shows the best results in terms of detecting domiciles, being able to detect 83\% of the domiciles present within the testing images. On the other hand, we carry out 90\% of correct classification using CCT for every category, which also requires a much smaller number of parameters compared to YOLOv8. Table \ref{tab: parameters} shows the total number of parameters for each model that we implement, including the YOLOv8, thus helping to provide a comparison between the computational efficiency of the models. The YOLOv8 model that we implement consists of approximately 25.9 million parameters while the CCT model consists of approximately 400 thousand parameters, which is fifty times lower than that of the number of parameters of YOLOv8. The Swin Transformer model and the EANet model consist of even lower parameters and the ViT model consists of approximately eleven million parameters, which is around half the number of parameters in YOLOv8. Except for EANet, the other three transformer-based architectures, which are CCT, ViT, and Swin Transformer show a relatively equal or better performance as their performance metric values for classification are over 80\% while having a very low number of parameters. For the CNN-based architectures, the number of parameters is quite higher than those of the transformer-based architectures, but still lower than that of YOLOv8. Only the custom ensemble model DCECNN that we implement has a higher number of parameters. Thus, we can infer from these observations and the observations from the results obtained by the models that transformer-based architectures like the CCT, Swin Transformer, and ViT are going to be a wiser choice compared to YOLOv8 in terms of categorizing humans and houses from the imagery of our dataset.

\subsection{Summary Findings}

Examining the results from our desired architectures, it is evident that transformer-based architectures outperform individual CNN-based architectures. Among these, the CCT demonstrates the best performance metrics while maintaining a low number of parameters, making it an excellent model for mobile device implementation. If we extend our proposed dataset AFSSA by collecting more images in the future, transformer-based architectures are likely to show even better performance. Given that transformer-based architectures, especially those used for computer vision, typically require a large amount of data, this outcome aligns with expectations. The self-attention layer of transformer-based models, such as Vision Transformer, lacks locality inductive bias - the assumption that image pixels are locally correlated and that their correlation maps are translation-invariant. Consequently, these models require more data to compensate for this lack of inductive bias.

Drones and image-processing models together have opened up numerous opportunities across various industries. In our work, drones can quickly assess damage in disaster-affected areas and locate survivors. Image processing aids rescue efforts by identifying trapped individuals or hazardous situations. Despite the significant potential, challenges remain, particularly the need for advanced onboard computing power. The CCT model, which has demonstrated the best performance with a very low number of parameters, stands out in terms of both performance and efficiency. Overall, a high accuracy, low number of parameters and low inference time make CCT a suitable model for implementation on mobile devices such as drones. This integration enhances the efficiency and effectiveness of drones in various applications, heralding a new era of possibilities.

\subsection{Limitations and Future Work}
\label{Limitations and future work}

As we work on flood imagery, certain limitations arise when creating our dataset. Most flood-prone South Asian countries are riverine, with numerous river channels and ponds scattered throughout rural areas. Consequently, some of the images collected for the `no flood' category contain these water bodies, introducing a slight bias. Models may incorrectly recognize these water bodies as flood zones, potentially altering the accuracy of our models by a small margin. Additionally, our proposed dataset consists solely of daytime flood imagery, meaning models trained on it will not be able to classify aerial flood images taken at night.

In our future work, we plan to enhance our dataset by expanding it with more images for each class, sourced from as many avenues as possible. We also aim to explore more architectures to improve classification accuracy and efficiency. Moreover, we plan to implement the classification of mixed classes. We hope that our study aids researchers in analyzing more computationally efficient methods to classify flood zones in South Asia from aerial imagery using our custom dataset, AFSSA.

\section{Conclusion}
\label{Conclusion}

The flood crisis in South Asian countries like India, Bangladesh, and Pakistan affects millions of inhabitants annually, with the numbers steadily increasing. Despite strengthened search and rescue initiatives, human casualties remain high as victims often cannot be reached in time. Our research aims to ease the search process by enhancing the efficiency of flood scene classification to improve rescue missions in these regions. We propose a dataset containing imagery from South Asia and use transformer-based models to train on this data, achieving high accuracy in classifying flood zones and houses within these areas. Notably, a fine-tuned Compact Convolutional Transformer provides an accuracy of 98.62\% with very few parameters, making it computationally efficient.

Our comparative analysis of different architectures and datasets offers insights into which types of architectures perform best in this context and explains why classification models are more effective than detection models. By enabling UAVs and aircraft to categorize houses, buildings, and people in flood zones, our approach allows rescue teams to quickly locate and respond to specific areas, reducing the need for extensive searches. Some potential research avenues include exploring the potential for real-time processing and integration into existing UAV platforms and expanding the dataset to include more diverse geographical locations and climatic conditions, which could further improve the model's generalizability and robustness. Moreover, the integration of other remote sensing technologies, such as LiDAR or Synthetic Aperture Radar (SAR), can be able to provide valuable complementary information for a more comprehensive understanding of flood scenes, especially in areas with poor visibility.

\section{Data Availability}

The dataset created and examined in this research can be provided by the corresponding author upon a reasonable request.

 \bibliographystyle{elsarticle-num} 
 \bibliography{Bibliography}

\end{document}